\RequirePackage{fix-cm}
\documentclass[smallextended]{svjour3}       
\smartqed  
\global\long\def\Kb{{\bf K}}%

\global\long\def\RR{{\bf \mathbb{R}}}

\global\long\def\Db{{\bf D}}%

\global\long\def\Bb{{\bf B}}%

\global\long\def\Ib{{\bf I}}%

\global\long\def\Mb{{\bf M}}%

\global\long\def\Xb{{\bf X}}%

\global\long\def\Yb{{\bf Y}}

\global\long\def\Ub{{\bf U}}

\global\long\def\Vb{{\bf V}}

\global\long\def\vsf{\mathsf{v}}

\global\long\def\xb{{\bf x}}

\global\long\def\yb{{\bf y}}

\global\long\def\zb{{\bf z}}%

\global\long\def\Sb{{\bf S}}%

\usepackage{amsfonts}
\usepackage{hyperref}
\hypersetup{
    colorlinks=true,
    linkcolor=black,
    filecolor=magenta,      
    urlcolor=blue,
}

\usepackage{microtype}
\usepackage{graphicx}
\usepackage{booktabs} 
\usepackage{amsmath}
\usepackage{algorithm, algpseudocode}
\usepackage{mathtools}
\usepackage{makecell}
\usepackage{subcaption}
\usepackage[moderate]{savetrees}  

\begin{document}

\title{Large Scale Tensor Regression using Kernels and Variational Inference
}
\subtitle{}


\author{Robert Hu \and Geoff K. Nicholls
        \and Dino Sejdinovic}


\institute{F. Author \at
              first address \\
              Tel.: +123-45-678910\\
              Fax: +123-45-678910\\
              \email{fauthor@example.com}           
           \and
           S. Author \at
              second address
}

\date{Received: date / Accepted: date}

\maketitle

\begin{abstract}
We outline an inherent weakness of tensor factorization models when latent factors are expressed as a function of side information and propose a novel method to mitigate this weakness. We coin our method \textit{Kernel Fried Tensor}(KFT) and present it as a large scale forecasting tool for high dimensional data. Our results show superior performance against \textit{LightGBM} and \textit{Field Aware Factorization Machines}(FFM), two algorithms with proven track records widely used in industrial forecasting. We also develop a variational inference framework for KFT and associate our forecasts with calibrated uncertainty estimates on three large scale datasets. Furthermore, KFT is empirically shown to be robust against uninformative side information in terms of constants and Gaussian noise.  
\end{abstract}

\section{Background and Problem}
\label{section:backg}
Large-scale prediction problems arising from the use of machine learning methods in industry are often formulated as an empirical risk minimization task coupled with a dataset and a model. In industrial applications, a high premium is often placed on calibrated uncertainty estimates around predictions. Nowadays, datasets often scale beyond the point where the standard data matrix formalism is practical, and this restricts the choice of models as they must give a computationally tractable analysis. In addition,  obtaining calibrated uncertainty estimates becomes challenging. In this paper we focus on tensor factorization models, a rich model class which provide tractability at scale. Tensor models are used in large scale prediction tasks ranging from bioinformatics to industrial forecasting; we work in the latter setting. However, while existing tensor models are versatile and scalable, they have a weakness: as we explain below, when we build a model of the latent factors in tensor factorization as a function of covariates\cite{Kim2014}\cite{Kim2016}, the model may be restricted and global parameter couplings are generated. These lead to reduction of tractability at scale. We give a new family of tensor models which admit side information without model restriction or loss of tractability.

\paragraph{Tensor Train decomposition}
Before we explain how side information restricts tensor model expressiveness, we set out the background. Consider the task of reconstructing the tensor $\Yb\in \mathbb{R}^{n_1 \times n_2 ... \times n_P}$. Many existing decomposition techniques \cite{Kolda09tensordecompositions} treat this problem. We focus on the Tensor Train (TT) decomposition \cite{doi:10.1137/090752286} as this generalises more readily than existing alternatives. \\ \newline The $n$-mode (matrix) product of a tensor $\mathcal X \in \mathbb{R}^{I_1 \times I_2 \times \ldots \times I_N}$ with a matrix $\mathbf U \in \mathbb{R}^{J \times I_n}$ is $\mathcal{X} \times_n \mathbf U$. This product is of size $I_1 \times \ldots I_{n-1} \times J \times I_{n+1} \times \ldots \times I_n$. Elementwise, the product is $$\left(\mathcal X \times_n \mathbf U \right)_{i_1\ldots i_{n-1}\ j\ i_{n+1} \ \ldots \ i_N} = \sum_{i_n=1}^{I_n}x_{i_1i_2\ldots i_N }\ u_{j i_n}.$$ We will consider the following notion of a mode product between tensor $\Xb \in \RR^{I_1 \times ... I_{N-1}\times I_{N}}$, applying the N-th mode product $\times_N$ with a tensor $\Ub \in \RR^{I_N\times K_1 \times K_2}$ gives $\Xb \times_N \Ub \in \RR^{I_1 \times ... I_{N-1}\times K_1 \times K_2}$. \\
\newline In TT, $\Yb$ is decomposed into $P$ latent tensors $\Vb_p\in \RR^{R_{p} \times n_p \times R_{p-1} }, p=1,...P$, with $\Vb_1 \in \RR^{n_1\times R_1}$. Here $R_p$ is the latent dimensionality for each factor in the decomposition, with $R_1=R_P=1$. Denote by $\times_{-1}  \Vb_p$ the mode product applied to the last dimension of $\Vb_p$. We seek $V_1,...,V_P$ so that
\begin{equation}
\begin{split}
    \Yb &\approx  \prod_{p=1}^P \times_{-1} \Vb_p\\
    y_{i_1...i_P} &\approx \sum_{r_0,...,r_P} \prod_{p=1}^P v_{r_{p}i_p r_{p-1}}.
\end{split}
\end{equation}
Suppose that, associated with dimension $p$, we have $c_p$-dimensional side information denoted $\Db^p \in \RR^{n_p \times c_p}$. For example, if $p$ is the dimension representing $n_p=10000$ different books, then the columns of $\Db^p\in \RR^{10000\times c_p}$ might contain the author of the book, page count etc. Similar to \cite{Kim2014}\cite{Kim2016}, side information is built into the second dimension of the latent tensor $\Vb_p\in \RR^{R_{p} \times c_p \times R_{p-1}}$ using the mode product $\Vb_p \times_{2} \Db^p$. For TT decomposition our approximation becomes

\begin{equation}
\begin{split}
    \Yb &\approx  \prod_{p=1}^P \times_{-1} (\Vb_p\times_{2} \Db^p)\\
    y_{i_1...i_P} &\approx \sum_{\substack{r_0,...,r_P\\i_1',...,i_P'}} \prod_{p=1}^P v_{r_{p}i_p'r_{p-1}}d_{i_p,i_p'}
\end{split}
\end{equation}
The above example illustrates the \textit{primal} setting where side information is applied directly. Similarly to \cite{Kim2016}, we also consider kernelized side information in the Reproducing Kernel Hilbert Space (RKHS) which we will refer to as the \textit{dual} setting.

\subsection{The Problem with adding side information to tensor factorization}

Consider a matrix factorization problem for $\Yb \in \RR^{N_1\times N_2}$ with unknown latent factors $\Ub\in \RR^{N_1\times R},\Vb\in \RR^{N_2\times R}$. We are approximating

\begin{equation}
\begin{split}
       \Yb &= 
\begin{bmatrix} 
    y_{11} & \dots & y_{1N_2} \\
    \vdots & \ddots & \\
    y_{N_11} &        & y_{N_1N_2} 
\end{bmatrix}
\approx \Ub\cdot \Vb^{\top} =  
\begin{bmatrix} 
    \sum_{r=1}^R u_{1r}v_{1r}  & \dots&\sum_{r=1}^R u_{1r}v_{N_2r} \\
    \vdots & \ddots & \\
    \sum_{r=1}^R u_{N_1r}v_{1r} &        & \sum_{r=1}^R u_{N_1r}v_{N_2r} 
\end{bmatrix}.
\end{split}
\end{equation}
If we update $u_{1r}$ in the approximation $y_{11}\approx\sum_{r=1}^R u_{1r}v_{1r}$, we change the approximation $y_{1,2}\approx\sum_{r=1}^R u_{1r}v_{2r}$ since they share the parameter $u_{1r}$. However, $y_{2,1}$ and $y_{2,2}$ remain unchanged. Parameters are coupled across rows and columns but not globally. This is the standard setup in latent factorization. \newline \\ Now consider an extreme case where we have $ \Db^1 = \Db^2= \mathbf{1}_{N_1\times N_1}$. We take our latent factors to be a linear function of available side information which leads $\Ub,\Vb$ to form $ \Db^1\Ub = 
\begin{bmatrix} 
    \sum_{i=1}^{N_1} u_{i1} & \dots & \sum_{i=1}^{N_1} u_{iR} \\
    \vdots & \ddots & \\
    \sum_{i=1}^{N_1} u_{i1} &        & \sum_{i=1}^{N_1} u_{iR}
\end{bmatrix}
$ and $\Db^2\Vb$ (similar form). It follows that 

\begin{equation}
\begin{split}
    &(\Db^1\Ub) \cdot (\Db^2\Vb)^{\top} = 
\begin{bmatrix} 
    \sum_{rij} u_{ir} v_{jr} & \dots & \sum_{rij} u_{ir} v_{jr} \\
    \vdots & \ddots & \\
    \sum_{rij} u_{ir} v_{jr} &        & \sum_{rij} u_{ir} v_{jr}
\end{bmatrix},
\end{split}
\end{equation}
is a constant matrix! We have lost all model flexibility as we are approximating $\Yb$ with a constant. Now consider a more realistic example with $\Db^1 =
\begin{bmatrix} 
    d_{11} & \dots & d_{1N_1} \\
    \vdots & \ddots & \\
    d_{N_11} &        & d_{N_1N_1} 
\end{bmatrix}$
and $\Db^2 =\begin{bmatrix} 
    z_{11} & \dots & z_{1N_2} \\
    \vdots & \ddots & \\
    z_{N_21} &        & z_{N_2N_2} 
\end{bmatrix}$. In this case

\begin{equation}
\begin{split}
    &(\Db^1\Ub) \cdot (\Db^2\Vb)^{\top} =  
\begin{bmatrix} 
    \sum_{rij} d_{1i}z_{1j}u_{ir} v_{jr} & \dots & \sum_{rij} d_{1i}z_{N_2j}u_{ir} v_{jr} \\
    \vdots & \ddots & \\
    \sum_{rij} d_{N_1i}z_{1j}u_{ir} v_{jr} &        &  \sum_{rij} d_{N_1i}z_{N_2j}u_{ir} v_{jr}
\end{bmatrix}.
\end{split}
\end{equation}
Again, $u_{ir}$ appears in all entries in our matrix approximation for $\Yb$. However this time changing $u_{ir}$ will not change all entries by the same amount but rather differently across {\it all} entries depending on the entries of $\Db^1$ and $\Db^2$. This connects all entries in the approximating matrix, introduces complex global variable dependence and makes fitting infeasible at large scale as the optimisation updates globally. The observation applies in both primal and dual representations. In the primal representation there is a restriction in expressiveness as the rank of our approximation falls off with the rank of the side information. In this setting near-colinearity is also a problem as it leads to unstable optimisation and factorisations which are very sensitive to noise.\\ \newline We see that when we add side information we may inadvertently restrict the expressiveness of our model. We formulate a new tensor model with a range and dependence structure unaffected by addition of side information in either the primal or dual setting. \\ \newline With the problem illustrated, we go on to answer the following questions:
\begin{enumerate}
    \item Is there a TT model class that avoids constraints and complex global dependencies arising from addition of side information but still makes full use of side information?
    \item Can we formulate and characterise this model class in both primal and dual space?
    \item Do models in this class compare favourably, for large scale prediction, to state-of-the-art models such as \textit{LightGBM} \cite{Ke2017LightGBMAH} and \textit{Field-Aware Factorization Machines} \cite{Juan:2016:FFM:2959100.2959134}?
    \item Can we work with these new models in a scalable Bayesian context with calibrated uncertainty estimates?
    \item What are the potential pitfalls of using these models? When are they appropriate to use?
\end{enumerate}

\section{Proposed Approach}

\subsection{Tensors and side information}
We seek a tensor model that benefits from additional side information while not forfeiting model flexibility. We introduce two strategies, which we call \textit{Weighted latent regression} (WLR) and \textit{Latent scaling} (LS).

\subsubsection{Weighted latent regression}
We now return to the previous setting but with additional latent tensors $\Ub'\in\RR^{N_1 \times R}$ and $\Vb'\in \RR^{N_2 \times R}$. 
We approximate $\Yb$ as: 

\begin{equation}
\begin{split}
    &\Yb \approx ((\Db^1\Ub) \circ \Ub') \cdot ((\Db^2\Vb) \circ \Vb')^{\top} = \\ &\tiny{
    \begin{bmatrix} 
    \sum_{rij}u_{1r}'v_{1r}' d_{1i}z_{1j}u_{ir} v_{jr} & \dots & \sum_{rij} u_{1r}'v_{N_2r}'d_{1i}z_{N_2j}u_{ir} v_{jr} \\
    \vdots & \ddots & \\
    \sum_{rij}u_{N_1r}'v_{1r}' d_{1i}z_{1j}u_{ir} v_{jr} &        &  \sum_{rij}u_{N_1r}'v_{N_2r}' d_{1i}z_{1j}u_{ir} v_{jr}
\end{bmatrix}}
\end{split}
\end{equation}
By taking the Hadamard product ($\circ$) with additional tensors $\Ub'$, $\Vb'$ we recover the model flexibility and dependence structure of vanilla matrix factorization, as $\Ub'$ and $\Vb'$ are independent of $\Ub$ and $\Vb$. Here, changing $u_{ir}$ would still imply a change for all entries by magnitudes defined by the side information, however we can calibrate these changes on a latent entrywise level by scaling each entry with $u_{ir}'v_{jr}'$. For any TT decomposition with an additional tensor $\Vb_p'$, the factorization becomes 

\begin{equation}
\begin{split}
    \Yb &\approx  \prod_{p=1}^P \times_{-1} ( \Vb_p' \circ (\Vb_p\times_{2} \Db^p))\\
    y_{i_1...i_P} &\approx \sum_{\substack{r_1,...,r_P\\i_1',...,i_P'\\i_1'',...,i_P''}} \prod_{p=1}^P v_{r_{p}i_p''r_{p-1}}'v_{r_{p}i_p'r_{p-1}}d_{i_p,i_p'}\delta_{i_p}(i_p'')
\end{split}
\end{equation}
where $\delta(\cdot)$ denotes Kronecker delta. We interpret this as weighting the regression terms $\sum_{i,j}d_{pi}z_{qj}u_{ir}v_{jr}$ over indices $p,q$  with $u_{pr}v_{qr}$ and then summing over latent indices $r$.

\subsubsection{Latent scaling}
An alternative computationally cheaper procedure would be to consider additional latent tensors $\Ub_1\in\RR^{N_1\times K},\Ub_2\in\RR^{N_1\times L},\Vb_1\in\RR^{N_2\times K},\Vb_2\in\RR^{N_2\times L}$. We would approximate 
\begin{equation}
    \begin{split}
            &\Yb \approx  (\Ub_1\Vb_1^{\top})\circ ((\Db^1\Ub) \cdot (\Db^2\Vb)^{\top}) + (\Ub_2\Vb_2^{\top})= \\ &\tiny{
    \begin{bmatrix} 
    \sum_{q}u_{1k}^{1}v_{1k}^{1} \sum_{rij} d_{1i}z_{1j}u_{ir} v_{jr}+\sum_{l}u_{1l}^{2}v_{1l}^{2} & \dots &  \\
    \vdots & \ddots & \\
    \sum_{q}u_{N_1k}^{1}v_{1k}^{1} \sum_{rij} d_{1i}z_{1j}u_{ir} v_{jr}+\sum_{l}u_{N_1l}^{2}v_{1l}^{2} &        
\end{bmatrix}}
    \end{split}
\end{equation}
We have similarly regained our original model flexibility where have introduced back independence for each term by scaling (adding a constant) for each regression term with a 'latent scale and bias' term. We generalize this to

\begin{equation}
\begin{split}
    &\Yb \approx  (\prod_{p=1}^P \times_{-1} \Vb_p^s)\circ(\prod_{p=1}^P \times_{-1} (\Vb_p\times_{2} \Db^p))+(\prod_{p=1}^P \times_{-1} \Vb_p^b)\\
    &y_{i_1...i_P} \approx \sum_{\substack{r^s_1,...,r^s_P}}\prod_{p=1}^Pv^s_{r_{p}i_pr_{p-1}} \sum_{\substack{r_1,...,r_P\\i_1',...,i_P'}} \prod_{p=1}^P v_{r_{p}i_p'r_{p-1}}d_{i_p,i_p'}+\sum_{\substack{r^b_1,...,r^b_P}}\prod_{p=1}^Pv^b_{r_{p}i_pr_{p-1}}
\end{split}
\end{equation}
The reader may conjecture that adding side information to tensors through a linear operation is counterproductive due to the restrictions it imposes on the approximation, and argue that our proposal of introducing additional tensors to increase model flexibility is futile when side information is likely to be marginally informative or potentially uninformative. As an example, return to the case of completely non-informative constant side information, $\Db^1=\mathbf{1}$, $\Db^2=\mathbf{1}$. In this corner case, both our proposed models reduce to regular matrix factorization: the side information regression term collapses to a constant, which in conjunction with the added terms reduces to regular tensor factorization without side information.  

\paragraph{A comment on identifiability}
 It should be noted that the proposed models need not be indentifiable.
 To see this, return to the scenario where side information is constant. The term $\Vb_p \times_2 \Db^p$ has constant rows equal to row sums of $\Vb_p$. Any transformation of $\Vb_p$ which preserves these row sums leaves the fit unchanged. However, in large-scale industrial forecasting we draw utility from good generalization performance in prediction, and parameter identifiability and interpretation is secondary.

\subsection{Primal regularization terms}

In weight space regression, the regularization terms given by the the squared Frobenius norm. The total regularization term would be written as

\begin{equation}
    \Lambda=\sum_{p} \lambda_{p} \Vert \Vb_p \Vert^2_F +\lambda_{p}' \Vert \Vb_p' \Vert^2_F 
\end{equation}

\subsection{RKHS and the Representer Theorem}
We extend our framework to also include the RKHS dual space formalism. The extension can intuitively be viewed as a tensorized version of Kernel Ridge Regression. Firstly consider side information $\Db^{p} = \{\xb_i^p\}_{i=1}^{N_p}, \xb_i^p \in \RR^{D_p}$ which are kernelized using a kernel function $k:\RR^{D_p}\times\RR^{D_p}\to\mathbb{R}$. Denote $k^{p}_{ij}=k(\xb_i^p,\xb_j^p) \in \RR$ and $\Kb^p=k(\Db^p,\Db^p)\in \RR^{N_p \times N_p}$. 
Consider a $\Vb \in \RR^{R_1\times n_1 ...\times n_Q\times R_2}$, where $Q<P$. Using the Representer theorem \cite{Schlkopf2001AGR} we can express  $ \prod_{q=1}^{Q}\Vb \times_{q+1} \Kb^{q}$ as a function in RKHS
\begin{equation}
    \vsf_{r_1,r_2}=\sum_{n_{1},...,n_q}v_{r_1n_1...n_qr_2}\prod_{q=1}^{Q} k_q(\cdot,\xb_{n_q}^{q}).
\end{equation}
Where $\vsf_{r_1,r_2}:\prod_{q=1}^Q \times \RR^{q} \to \RR$ and $\vsf_{r_1,r_2} \in \mathcal{H}$, which denotes the RKHS with respect to the kernels $\prod_{q}^Q \times k_q$.
\subsection{Dual space regularization term for WLR}

Consider applying another tensor with the same shape as $\Vb'$ through element wise product to robustify $\Vb$, i.e. $\Vb' \circ \prod_{q=1}^{Q}\Vb \times_{q+1} \Kb^{q}$. Then we have

\begin{equation}
\begin{split}
        \vsf_{r_1r_2}&= \left(\sum_{n_{1}',...,n_q'}v_{r_1n_1'...n_q'r_2}'\prod_{q=1}^{Q}\delta_{\xb_{n_q'}^{q}}(\cdot)\right)\cdot \left(\sum_{n_{1},...,n_q}v_{r_1n_1...n_qr_2}\prod_{q=1}^{Q} k(\cdot,\xb_{n_q}^{q})\right)\\&=\sum_{\substack{n_1,...,n_q \\n_1',...,n_q}}v_{r_1n_1...n_qr_2}v_{r_1n_1'...n_q'r_2}'\prod_{q=1}^{Q}\delta_{\xb_{n_q'}^{Q}}(\cdot)\prod_{q=1}^{Q}k(\cdot,\xb_{n_q}^{q})
\end{split}
\end{equation} where the regularization term for $\vsf_{r_1,r_2}$ is given by:
\begin{equation}
\begin{split}
        \Lambda&=\sum_p \lambda_p \sum_{r_1,r_2} \langle\vsf_{r_1,r_2},\vsf_{r_1,r_2} \rangle_{\mathcal{H}}\\&=\sum_p \lambda_p\Bigg((\prod_{q=1}^Q \Vb\times_{q+1}\Kb^q) \circ \left( (\prod_{q=1}^Q(\Vb'\circ\Vb')\times_{q+1} \mathbf{1}_{n_q\times n_q})\circ \Vb  \right)\Bigg)_{++}
\end{split}
\end{equation}
and $(\cdot)_{++}$ means summing all elements. 

\subsection{Dual space regularization term for LS}

For LS models, the regularization term is calculated as 

\begin{equation}
    \Lambda = \sum_p \lambda_p \Bigg[\Vert \Vb_p^s\Vert^2_F + \sum_{r} \langle \vsf_r,\vsf_r \rangle_{\mathcal{H}} + \Vert \Vb_p^b\Vert^2_F\Bigg]
\end{equation}
where $\vsf_{r_1,r_2}= \sum_{n_{1},...,n_q}v_{r_{p}n_1...n_qr_{p-1}}\prod_{q=1}^{Q} k(\cdot,\xb_{n_q}^{Q})$ and \\ $\sum_{r}\langle \vsf_r,\vsf_r \rangle_{\mathcal{H}} = \Bigg((\prod_{q=1}^Q \Vb\times_{q+1}\Kb^q) \circ \Vb \Bigg)_{++}$.

\subsection{Scaling with Random Fourier Features} 
To make tensors with kernelized side information scalable, we rely on a Random Fourier Feature \cite{10.5555/2981562.2981710} (RFFs) approximation of the true kernels. RFFs approximate a translation-invariant kernel function $k$ using Monte Carlo:
\vspace{-0.3pt}
\begin{equation}
    \hat{k}(\xb,\yb) =
\frac{2}{I}\sum_{i=1}^{I/2} \big[\cos(\omega_i^T \xb)\cos(\omega_i^T \yb) + \sin (\omega_i^T \xb) \sin (\omega_i^T \yb)\big] 
\end{equation}
where $\omega_i$ are frequencies drawn from a normalized non-negative spectral measure $\Lambda$ of kernel $k$. Our primary goal in using RFFs is to create a memory efficient, yet expressive method. Thus, we write

\begin{equation}
\hat{k}\left(\xb_{i}^p,\xb_j^p \right) \approx \phi\left(\xb_{i}^p\right)^{\top}\phi\left(\xb_{j}^p\right)
\end{equation}
with explicit feature map
$\phi:\mathbb{R}^{D_p}\to\mathbb{R}^{I}$, and $I\ll N_p$. In the case of RFFs, 
$$\phi(\cdot)^{\top}=[\cos(\omega_1^T \cdot),\ldots,\cos(\omega_{I/2}^T \cdot),\sin(\omega_1^T \cdot),\ldots,\sin(\omega_{I/2}^T \cdot)].$$ This feature map can be applied in the primal space setting as a computationally cheap alternative to the RKHS dual setting. \newline \\
A drawback of tensors with kernelized side information is the $\mathcal{O}(N_p^2)$ memory growth of kernel matrices. If one of the dimensions has a large $N_p$ in the dual space setting, we approximate large kernels $\Kb^p$ with

\begin{equation}
    \Vb \times_{p+1} \Phi \times_{p+1} \Phi^{\top} \approx \Vb \times_{p+1} \Kb^p,   
\end{equation}
where $\Phi=\phi(\Sb^p)\in \RR^{N_p\times I}$. To see that this is a valid approximation, an element $\bar{v}_{i_1...i_p}$ in $\Vb\times_p \Kb^p$ is given by $\bar{v}_{i_1...i_p} = \sum_{i_p'=1}^{N_p}v_{i_1...i_p'}k(\xb_{i_p}^p,\xb_{i_p'}^p)$. Using RFFs we have

\begin{equation}
    \bar{v}_{i_1...i_p} = \sum_{i_p'=1}^{N_p}\sum_{c=1}^I v_{i_1...i_p'}\Phi_{i_p'c}  \Phi_{i_p c}=\sum_{i_p'=1}^{N_p}v_{i_1...i_p'}\phi(\xb_{i_p'}^p)^{\top}  \phi(\xb_{i_p}^p).
\end{equation}
KFT with RFFs now becomes:
\begin{equation}
\begin{split}
        &\vsf_{r_p} =\sum_{\substack{n_1,...,n_q \\n_1',...,n_q}}v_{r_{p}n_1...n_qr_{p-1}}v_{r_{p}n_1'...n_q'r_{p-1}}'\prod_{q=1}^{Q}\delta_{\xb_{n_q'}^{Q}}(\cdot)\prod_{q=1}^{Q}\phi(\xb_{n_q}^{Q})\phi(\cdot)
\end{split}
\end{equation}
With the regularization term

\begin{equation}
\begin{split}
    &\sum_{r}\langle \vsf_r,\vsf_r \rangle_{\mathcal{H}}=\Bigg((\prod_{q=1}^Q \Vb\times_{q+1} \Phi_{q}\times_q \Phi_q^{\top} ) \circ \left( (\prod_{q=1}^Q(\Vb'\circ\Vb')\times_q \mathbf{1}_{n_q\times n_q})\circ \Vb  \right)\Bigg)_{++}.
\end{split}
\end{equation}
For a derivation, please refer to the appendix.

\subsection{Kernel Fried Tensor}

Having established our new model, we coin it Kernel Fried Tensor (KFT). Given an empirical risk minimization objective $\mathcal{L}(\Yb,\Tilde{\Yb})$ for predictions $\Tilde{\Yb}$ the full objective is 

\begin{equation}
    \min_{\text{wrt } \Vb_p,\Vb_p',\Theta_p} \mathcal{L}(\Yb,\Tilde{\Yb}) + \Lambda(\Vb_p,\Vb_p',\Theta_p)
\end{equation}
where $\Vb_p,\Vb_p',\Theta_p$ are parameters of the model and $\Theta_p$ are kernel parameters if we use the RKHS dual formulation.


\subsubsection{Training Procedure}
As our proposed model involves mutually dependent components with non-zero mixed partial derivatives, optimizing them jointly with a first order solver is inappropriate as mixed partial derivatives will not be considered during each gradient step. Inspired by the EM-algorithm \cite{10.2307/2984875}, we summarize our training procedure in algorithm \ref{alg:EM_train}.
\begin{algorithm}
\begin{algorithmic}
\Require $\Vb_p,\Vb_p',\Kb^p,\Theta_p$, Data, learning rate $\alpha_{\text{lr}}$
\For{ epochs}
\For{parameters $\mathcal{P}$ in $\{\Vb_p,\Vb_p',\Theta_p\}$}
\State Fix all parameters $\{\Vb_p,\Vb_p',\Theta_p\}$ except $\mathcal{P}$
\For{batch in data}
\State Calculate $\mathcal{L}(\Yb,\Tilde{\Yb}) + \Lambda(\Vb_p,\Vb_p',\Theta_p)
$
\State Update $\mathcal{P}$ as  $\mathcal{P} = \mathcal{P} - \alpha_{\text{lr}}\frac{\partial(\mathcal{L}+\Lambda)}{\partial \mathcal{P}}$
\EndFor
\EndFor
\EndFor
\end{algorithmic}
\caption{Tensor training procedure}
\label{alg:EM_train}
\end{algorithm}
By updating each parameter group sequentially and independently, we eliminate the effect of mixed partials. For a more detailed motivation, see the Appendix.

\subsubsection{Joint features}

From a statistical point of view, we are assuming that each of our latent tensors $\Vb_p$ factorizes $\Yb$ into $P$ independent components with prior distribution corresponding to $\mathcal{N}(\vsf_{r_{p-1}r_p}^{p} |0,\Kb^p)$, where $\vsf_{r_{p-1}r_p}^{p}\in\RR^{n_p}$ is the $r_{p-1},r_p$ cell selected from $\Vb_p$. We can enrich our approximation by jointly modelling some dimensions $p$ by choosing some $\Vb_p\in\RR^{R_{p+1}\times n_{p+1}\times n_{p} \times R_{p-1  }}$. If we denote this dimension $p$ by $p'$ we have that
\begin{equation}
    \begin{split}
            &\Yb \approx \prod_{p=1}^{p'-1} \times_{-1} (\Vb_p\times_{2} \Kb^p)  \times_{-1}(\Vb_{p'}\times_3 \Kb^{p'} \times_{2} \Kb^{p'+1}) \prod_{p=p'+2}^P \times_{-1}(\Vb_p\times_{2} \Kb^p)\\&y_{i_1...i_P}\approx \sum_{\substack{r_1,...,r_P\\i_1',...,i_P'}} \prod_{p=1}^{p'-1} v_{r_{p}i_p'r_{p-1}}k(\xb_{i_p}^p,\xb_{i_p'}^p)\cdot v_{r_{p}i_{p'}'i_{p'+1}'r_{p-1}}k(\xb_{i_{p'}}^{p'},\xb_{i_{p'}'}^{p'})k(\xb_{i_{p'+1}}^{p'+1},\xb_{i_{p'+1}'}^{p'+1})\\&\prod_{p=1}^{p'+2} v_{r_{p}i_p'r_{p-1}}k(\xb_{i_p}^p,\xb_{i_p'}^p)
    \end{split}
\end{equation}
and the prior would instead be given as $\mathcal{N}(\text{vec}(\vsf_{r_{p'-1}r_{p'+1}}^{p'}) |0,\Kb^{p'}\otimes\Kb^{p'+1})$. Here $\text{vec}(\vsf_{r_{p'-1}r_{p'+1}}^{p'}) \in \RR^{n_{p'}n_{p'+1}}$ and $\text{vec}(\cdot)$ means flattening a tensor to a vector. The cell selected from $\Vb_{p'}$ now has a dependency between dimensions $p'$ and $p'+1$. We refer to a one dimensional factorization component of TT as a \textit{TT-core} and a multi dimensional factorization component as a \textit{joint TT-core}.






\subsection{Bayesian Inference}

To put KFT in a probabilistic framework we turn to Bayesian inference. We assume a Gaussian conditional likelihood for an observation $y_{i_1,...,i_P}$ with inspiration from \cite{pmlr-v28-gonen13a}\cite{Kim2014}. For WLR we have that

\begin{equation}
\begin{split}
        &p(y_{i_1,...,i_P}|\Vb_1,...,\Vb_P,\Vb_1',...,\Vb_P') =\\& \mathcal{N}\Bigg(y_{i_1,...,i_P}|\sum_{\substack{r_1,...,r_P\\i_1',...,i_P'\\i_1'',...,i_P''}} \prod_{p=1}^P v_{r_{p}i_p''r_{p-1}}'v_{r_{p}i_p'r_{p-1}}k(\xb_{i_p}^p,\xb_{i_p'}^p)\delta_{\xb_{i_p}^p}(\xb_{i_p''}^p),\sigma_y^2\Bigg).
\end{split}
\end{equation}
The corresponding objective for LS is

\begin{equation}
\begin{split}
    &p(y_{i_1,...,i_P}|\Vb_1,...,\Vb_P,\Vb_1',...,\Vb_P') =\\&
    \mathcal{N}\Bigg(y_{i_1...i_P}| \sum_{\substack{r^s_1,...,r^s_P}}\prod_{p=1}^Pv^s_{r_{p}i_pr_{p-1}} \sum_{\substack{r_1,...,r_P\\i_1',...,i_P'}} \prod_{p=1}^P v_{r_{p}i_p'r_{p-1}}k(\xb_{i_p}^p,\xb_{i_p'}^p)+\sum_{\substack{r^b_1,...,r^b_P}}\prod_{p=1}^Pv^b_{r_{p}i_pr_{p-1}},\sigma_y^2\Bigg)
\end{split}
\end{equation}
where $\sigma_y^2$ is a scalar hyperparameter. 




\subsection{Variational Approximation}
Our goal is to maximize the posterior distribution $p(\Vb_p,\Vb_p'|\Yb)$ which is intractable as the likelihood $p(\Yb)= \int p(\Yb|\Vb_1,...,\Vb_P,\Vb_1',...,\Vb_P')p(\Vb_1)...p(\Vb_P)\\p(\Vb_1')...p(\Vb_P')d\Vb_{1,...,P}d\Vb_{1,...,P}'$ does not have a closed form solution due to the product of Gaussians. Instead we use variational approximations for $\Vb_p,\Vb_p'$ by parametrizing distributions of the Gaussian family and optimize the Evidence Lower BOund(ELBO)

\begin{equation}
\begin{split}
         &\mathcal{L}(y_{i_1,...,i_P},\Vb_1,...,\Vb_p,\Vb_1',...,\Vb_p')= \mathbb{E}_{q}[\log{p(y_{i_1,...,i_P}|\Vb_1,...,\Vb_p,\Vb_1',...,\Vb_p')}]-
        \\&D_{\text{KL}}(q(\Vb_1,...,\Vb_p,\Vb_1',...,\Vb_p'|y_{i_1,...,i_P})\Vert p(\Vb_1,...,\Vb_p,\Vb_1',...,\Vb_p'))
\end{split}
\end{equation}
In our framework we consider the univariate Gaussian and multivariate Gaussian as variational approximations with corresponding priors where $\sigma_y^2$ is interpreted to control the weight of the reconstruction term against the KL-term.
\subsubsection{Univariate VI}

\paragraph{Univariate KL}
For the case of univariate normal priors, we calculate the KL divergence as
\begin{equation}
\begin{split}
{\displaystyle D_{\mathrm {KL} }(\mathcal{N}_q\,\|\,\mathcal{N}_p)={\frac {(\mu _{q}-\mu _{p})^{2}}{2\sigma _{p}^{2}}}+{\frac {1}{2}}\left({\frac {\sigma _{q}^{2}}{\sigma _{p}^{2}}}-1-\ln {\frac {\sigma _{q}^{2}}{\sigma _{p}^{2}}}\right)}
\end{split}
\end{equation}
where $\mu_q,\mu_p$ and $\sigma_q^2,\sigma_p^2$ are the mean and variance for the prior and variational approximation respectively, where $\mu_p,\sigma_p^2$ are chosen a priori.

\paragraph{Model}
For a univariate Gaussian variational approximation we assume the following prior structure
\begin{equation}
    \begin{aligned}
        p(\Vb_p')= \prod_{\substack{r_p,i_p}}\mathcal{N}(v_{r_{p}i_pr_{p-1}}'|\mu_p',\sigma_p'^2)\\
        p(\Vb_p) = \prod_{r_p,i_p}\mathcal{N}(v_{r_{p}i_pr_{p-1}}|\mu_p,\sigma_p^2)
    \end{aligned}
\end{equation}
with corresponding univariate meanfield approximation

\begin{equation}
    \begin{aligned}
        q(\Vb_p')= \prod_{\substack{r_p,i_p}}\mathcal{N}(v_{r_{p}i_pr_{p-1}}'|\mu_{r_pi_pr_{p-1}}',\sigma_{r_pi_pr_{p-1}'}'^2)\\
        q(\Vb_p) = \prod_{r_p,i_p}\mathcal{N}(v_{r_{p}i_pr_{p-1}}|\mu_{r_pi_pr_{p-1}},\sigma_{r_pi_pr_{p-1}}^2).
    \end{aligned}
\end{equation}
We take $\mu_p',\sigma_p'^2,\mu_p,\sigma_p^2$ to be hyperparameters.

\paragraph{Weighted latent regression reconstruction term}

For the case of Weighted latent regression, we express the reconstruction term as

\begin{equation}
    \begin{split}
        &\mathbb{E}_{q}[\log{p(y_{i_1,...,i_P}|\Vb_1,...,\Vb_p,\Vb_1',...,\Vb_p')}] \propto \\& \frac{1}{\sigma_y^2} \Bigg[\Yb^2-2\Yb \circ \Bigg( \prod_{p=1}^P \times_{-1} ( \Mb_p' \circ (\Mb_p\times_{2} \Kb^p))\Bigg) + \Bigg(\prod_p^P \times_{-1}   \Mb_p' \circ (\Mb_p\times_{2} \Kb^p)\Bigg)^2 \\&+ \Bigg(\prod_p^P \times_{-1}\Sigma_p' \circ (\Sigma_p\times_{2} (\Kb^p)^2)\Bigg)+\Bigg(\prod_p^P \times_{-1}(\Sigma_p' \circ(\Mb_p\times_2 \Kb^p)^2) \Bigg)\\&+\Bigg(\prod_p^P \times_{-1} \Mb_p'^2\circ (\Sigma_p\times_{2} (\Kb^p)^2)\Bigg) \Bigg]
    \end{split}
\end{equation}
where $\Mb_p',\Mb_p$ and $\Sigma_p',\Sigma_p$ correspond to the tensors containing the variational parameters $\mu_{r_{p}i_pr_{p-1}}'$, $\mu_{r_{p}i_pr_{p-1}}$ and $\sigma_{r_{p}i_pr_{p-1}}'^2$, $\sigma_{r_{p}i_pr_{p-1}}^2$ respectively. For the case of RFF's, we approximate $ \Sigma_p \times_2 (\Kb^p)^2 \approx \Sigma_p \times_2 (\Phi_p \bullet \Phi_p)^{\top} \times_2 (\Phi_p \bullet \Phi_p) $, where $\bullet$ is the transposed Khatri-Rao product. It should further be noted that any square term means element wise squaring. We provide a derivation in the appendix.

\paragraph{Latent scaling reconstruction term}
For the case of Weighted latent regression, we express the reconstruction term as

\begin{equation}
\begin{split}
     &\mathbb{E}_{q}[\log{p(y_{i_1,...,i_P}|\Vb_1,...,\Vb_p,\Vb_1',...,\Vb_p')}] \propto \\&\frac{1}{\sigma_y^2} \Bigg[\Yb^2 -2\Yb\circ\Bigg(\prod_{p=1}^P (\Mb_p^s \circ (\Mb_p \times_2 \Kb^p) + \Mb_p^b) \Bigg)
    \\&+\Bigg((\prod_{p=1}^P \times_{-1} \Mb_p^s)^2+(\prod_{p=1}^P \times_{-1} \Sigma_p^s) \Bigg)\circ\Bigg((\prod_p^P \times_{-1} \Mb_p\times_{2} \Kb^p)^2+ \prod_{p=1}^P \times_{-1} \Sigma_p\times_{2} (\Kb^p)^2\Bigg)\\&+2(\prod_{p=1}^P \times_{-1} \Mb_p^s)\circ(\prod_{p=1}^P \times_{-1} \Mb_p^b)\circ(\prod_{p=1}^P \times_{-1} \Mb_p \times_2 \Kb^p) \\&+\Bigg((\prod_{p=1}^P \times_{-1} \Mb_p^b)^2+(\prod_{p=1}^P \times_{-1} \Sigma_p^b) \Bigg)\Bigg].
\end{split}
\end{equation} 
For details, see the appendix.
\subsubsection{Multivariate VI}

\paragraph{Multivariate KL}

The KL divergence between a multivariate normal prior $p$ and a variational approximation given by $q$:

\begin{equation}
\begin{split}
   &D_{\text{KL}}({\mathcal {N}}_{q}\parallel {\mathcal {N}}_{p})=\\&{\frac {1}{2}}\left[\operatorname {tr} \Bigg(\Sigma_{p}^{-1}\Sigma_{q}\right)+(\mu _{p}-\mu _{q})^{\mathsf {T}}\Sigma_{p}^{-1}(\mu _{p}-\mu _{q})-k+\ln \left({\frac {\det \Sigma_{p}}{\det \Sigma_{q}}}\right)\Bigg]
   \end{split}
\end{equation}
Where $\mu_p,\mu_q$ and $\Sigma_p,\Sigma_q$ are the mean and covariance for the prior and variational respectively. We take $\Sigma_p = \Kb^{-1}$, where $\Kb^{-1}$ is the kernel covariance matrix. The inverse $\Kb^{-1}$ is inspired by g-prior \cite{Zellner1986OnAP}. Another benefit of using the inverse is that is simplifies calculations, since we now avoid inverting a dense square matrix in the KL-term. Similar to the univariate case we choose $\mu_p$ a priori, although here it becomes a constant tensor rather than a constant scalar.  

\paragraph{Model}
For the multivariate case we consider the following priors

\begin{equation}
    \begin{split}
        &p(\Vb_p')= \prod_{\substack{r_p,i_p}}\mathcal{N}(v_{r_{p}i_pr_{p-1}}'|\mu_p',\sigma_p'^2)\\
            &p(\Vb_p) = \prod_{r_p}\mathcal{N}(\text{vec}(\vsf_{r_p})|\mu_p,\prod_{q=1}^{Q_p}\otimes(\Kb^{q})^{-1})
    \end{split}
\end{equation}
Where $Q_p$ is the number of dimensions jointly modeled in each TT-core. For the variational approximations we have

\begin{equation}
    \begin{split}
        &q(\Vb_p')= \prod_{\substack{r_p,i_p}}\mathcal{N}(v_{r_{p}i_pr_{p-1}}'|\mu_{r_pi_pr_{p-1}}',\sigma_{r_pi_pr_{p-1}}'^2)\\&q(\Vb_p) = \prod_{r_p,i_p}\mathcal{N}(\text{vec}(\vsf_{r_p})|\mu_{r_pi_pr_{p-1}},\prod_{q=1}^{Q_p}\otimes(\Bb_{q}\Bb_{q}^{\top}))
    \end{split}
\end{equation}
We take $\mu_p',\sigma_p'^2,\mu_p$ to be hyperparameters.

\paragraph{Sampling and parametrization}

Calculating $\prod_{q=1}^{Q_i} \otimes \Bb_{q}\Bb_{q}^{\top}$ directly will yield a covariance matrix that is prohibitively large. To sample from $q(\Vb_p)$ we exploit that positive definite matrices $A$ and $B$ with their Cholesky decompositions $L_A$ and $L_B$ have the following property

\begin{equation}
\label{hacK_1}
    A\otimes B = (L_AL_A^{\top})\otimes (L_BL_B^{\top}) = (L_A\otimes L_B) (L_A\otimes L_B)^{\top}.
\end{equation}
together with the fact that

\begin{equation}
\label{hack_2}
     \prod_{i=1}^{N} \Xb \times_{i+1} A_{i} =  \left(\prod_{i=1}^N \otimes A_i\right) \cdot \text{vec}(\Xb),
\end{equation}
where $\text{vec}(\Xb)\in\RR^{\prod_{i}^N I_i\times R}$. We would then draw a sample $\mathbf{b}\sim q(\Vb)$ as 

\begin{equation}
    \mathbf{b} = \mu_{r_p} + \prod_{q=1}^{Q_i} \Tilde{\zb} \times_{q+1} \Bb_q
\end{equation}
where $\Tilde{\zb}\sim \mathcal{N}(0,\mathbf{I}_{\prod_{p=1}^{P}n_p})$ is reshaped into $\Tilde{\zb}\in\RR^{\prod_{q=1}\times n_q}$. Taking inspiration from \cite{ong2017Gaussian}, we take $\Bb_q = \textbf{ltri}(B_qB_q^{\top})+D_q$, where $B_q\in \RR^{n_q\times  r}$, $D_q$ to be a diagonal matrix and $\textbf{ltri}$ denotes taking the lower triangular component of a square matrix including the diagonal. We choose this parametrization for a linear time-complexity calculation of the determinant in the KL-term. In the RFF case, we take $\Bb_q=B_q$ and estimate the covariance as $\Bb_q \Bb_q^{\top} + D_q^2$
\paragraph{Weighted latent regression reconstruction term} We similarly to the univariate case express the reconstruction term as

 \begin{equation}
    \begin{split}
        &\mathbb{E}_{q}[\log{p(y_{i_1,...,i_P}|\Vb_1,...,\Vb_p,\Vb_1',...,\Vb_p')}] \propto \\& \frac{1}{\sigma_y^2} \Bigg(\Yb^2-2\Yb \circ \Bigg( \prod_{p=1}^P \times_{-1} ( \Mb_p' \circ (\Mb_p\times_{2} \Kb^p))\Bigg) + \Bigg(\prod_p^P \times_{-1}   \Mb_p' \circ (\Mb_p\times_{2} \Kb^p)\Bigg)^2 \\&+ \Bigg(\prod_{p}^P\times_{-1}\Sigma_p'\circ ( \mathbf{1}\times_2  \Big(\Ib\cdot(\Kb^p \cdot \Bb_p)^2\cdot \bar{1})\Big) \Bigg)+\Bigg(\prod_p^P \times_{-1}(\Sigma_p' \circ(\Mb_p\times_2 \Kb^p)^2) \Bigg)\\&+ \Bigg(\prod_{p}^P\times_{-1} \Mb_p'^2\circ ( \mathbf{1}\times_2 \Big(\Ib\cdot(\Kb^p \cdot \Bb_p)^2\cdot \bar{1})\Big) \Bigg)
    \end{split}
\end{equation}
where $\Sigma_p =  \Bb_p  \Bb_p^T$, $\mathbf{1}$ denotes a constant one tensor with the same dimensions as $\Sigma_p'$, $\hat{1}\in \RR^{1\times R}$ where $R$ is the column dimension of $\Bb_p$, $\Ib$ the identity matrix and $\Sigma_p'$ is the same as in the univariate case.  For RFF's we have that

\begin{equation}
    (\Kb^p \cdot \Bb_p)^2\cdot \bar{1}\approx ((\Phi_p\cdot\Phi_p^{\top}) \cdot \Bb_p)^2 \cdot \bar{1} + \text{vec}(D_p^2)= (\Phi_p\cdot(\Phi_p^{\top} \cdot \Bb_p))^2\cdot \bar{1}+ \text{vec}(D_p^2).
\end{equation}

\paragraph{Latent Scaling}
The latent scaling version has the following expression

\begin{equation}
\begin{split}
     &\mathbb{E}_{q}[\log{p(y_{i_1,...,i_P}|\Vb_1,...,\Vb_p,\Vb_1',...,\Vb_p')}] \propto \\&\frac{1}{\sigma_y^2} \Bigg[\Yb^2 -2\Yb\circ\Bigg(\prod_{p=1}^P (\Mb_p^s \circ (\Mb_p \times_2 \Kb^p) + \Mb_p^b) \Bigg)+\Bigg((\prod_{p=1}^P \times_{-1} \Mb_p^s)^2+(\prod_{p=1}^P \times_{-1} \Sigma_p^s) \Bigg)\\&\circ\Bigg((\prod_p^P \times_{-1} \Mb_p\times_{2} \Kb^p)^2+ \prod_{p=1}^P \times_{-1} (\mathbf{1}\times_2\Big(\Ib\cdot(\Kb^p \cdot \Bb_p)^2\cdot \bar{1})\Big)\Bigg)\\&+2(\prod_{p=1}^P \times_{-1} \Mb_p^s)\circ(\prod_{p=1}^P \times_{-1} \Mb_p^b)\circ(\prod_{p=1}^P \times_{-1} \Mb_p \times_2 \Kb^p)+\Bigg((\prod_{p=1}^P \times_{-1} \Mb_p^b)^2\\&+(\prod_{p=1}^P \times_{-1} \Sigma_p^b) \Bigg) \Bigg].
\end{split}
\end{equation}
For details, see the appendix.

\paragraph{RFFs and KL divergence}
Using $(\Phi_p\Phi_p^{\top})^{-1}\approx (\Kb^{p})^{-1}$ as our prior covariance, we observe that the KL-term presents computational difficulties as a naive approach would require storing $(\Kb^{p})^{-1}\in \RR^{n_p\times n_p}$ in memory. Assuming we take $\Sigma_p = BB^{\top}, B \in \RR^{n_p \times R}$, we can manage the first term by using the equivalence

\begin{equation}
    (A \bullet B)\cdot (A \bullet B)^{\top} = (AA^{\top})\circ (BB^{\top}).
\end{equation}
Consequently we have that

\begin{equation}
\begin{split}
        &\text{tr}(\Sigma_{p}^{-1}\Sigma_q)=\bigg(\underbrace{{(\Phi\Phi^{\top})}}_{\text{Using }(\Kb^p)^{-1}} \circ  BB^{\top}\bigg)_{++} =\bigg( (\Phi \bullet B)\cdot (\Phi \bullet B)^{\top}\bigg)_{++} \\& = \bigg( (\Phi \bullet B)^{\top}\cdot (\Phi \bullet B)\bigg)_{++}.
\end{split}
\end{equation}
We can calculate the second term using \eqref{hacK_1} and \eqref{hack_2}. For the third term we remember Weinstein-Aronszajn's identity 

\begin{equation}
   \det(I_{m}+AB)=\det(I_{n}+BA) 
\end{equation}
 where $A\in\RR^{m\times n},B\in\RR^{n\times m}$ and $AB$ is trace class. If we were to take our prior covariance matrix to be $\Sigma_p = (\Phi_p\Phi_p^{\top} + I_{n_p} )^{-1}\approx (\Kb^{p} + I_{n_p})^{-1}$ and our posterior covariance matrix to be approximated as $\Sigma_q = BB^{\top}+I_{n_p}$, we could use Weinstein-Aronszajn's identity to calculate the third log term in a computationally efficient manner. \newline \\ From a statistical perspective, adding a diagonal to the covariance matrix implies regularizing it by increasing the diagonal variance terms. Taking inspiration from \cite{kim2017scaling}, we can further choose the magnitude $\sigma$ of the regularization

\begin{align*}
\det(\sigma^2 I +\Phi\Phi^\top) &=(\sigma^2)^n \det(I+\sigma^{-2}\Phi\Phi^\top) \\
&= (\sigma^2)^n \det(I+\sigma^{-2}\Phi^\top\Phi) \\
&= (\sigma^2)^{n-m} \det(\sigma^2 I+\Phi^\top\Phi) \\
\end{align*}
The KL expression then becomes

\begin{equation}
\begin{split}
   &D_{\text{KL}}({\mathcal {N}}_{q}\parallel {\mathcal {N}}_{p})={\frac {1}{2}}\Bigg[\left((\Phi \bullet B)^{\top}\cdot (\Phi \bullet B)\right)_{++}+\sigma_p^2(B\circ B)_{++}+\sigma_q^2 \Phi\circ\Phi)_{++}+\sigma_q^2\sigma_p^2N\\&  +(\mu _{p}-\mu _{q})^{\mathsf {T}}(\Phi\Phi^{\top}+\sigma_p^2I)(\mu _{p}-\mu _{q})-k+\ln \left({\frac {|(\sigma_p^2)^{N-I}\det (\Phi^{\top}\Phi+\sigma_p^2I)|^{-1}}{|(\sigma_q^2)^{N-I}\det(B^{\top}B+\sigma_q^2I)|}}\right)\Bigg].
   \end{split}
\end{equation}





\subsubsection{Optimization Scheme}
Similarly to the frequentist case, we use an EM inspired optimization strategy in \ref{alg:bayes_EM_train}. The main idea is to find the modes and variance parameters of our variational approximation in a mutually exclusive sequential order. Informally, we first find the modes and then the uncertainty. Similarly to the frequentist case, the reconstruction term of the ELBO has terms that both contain $\Sigma_p$ and $\Mb_p'$ which motivates the EM inspired approach.

\begin{algorithm}
\begin{algorithmic}
\Require $\Mb_p,\Mb_p',\Sigma_p,\Sigma_p',\Kb^p,\Theta_p$, Data, learning rate $\alpha_{\text{lr}}$
\For{parameter group $\mathcal{Z}$ in $\Big\{\{\Mb_p,\Mb_p',\Theta_p\},\{\Sigma_p,\Sigma_p'\}\Big\}$}
\For{ epochs}
\For{parameters $\mathcal{P}$ in $\mathcal{Z}$}
\State Fix all parameters in $\mathcal{Z}$ except $\mathcal{P}$
\For{batch in data}
\State Calculate ELBO $\mathcal{L}(y_{i_1,...,i_P},\Vb_1,...,\Vb_p,\Vb_1',...,\Vb_p')$
\State Update $\mathcal{P}$ as  $\mathcal{P} = \mathcal{P} - \alpha_{\text{lr}}\frac{\partial\mathcal{L}}{\partial \mathcal{P}}$
\EndFor
\EndFor
\EndFor
\EndFor
\end{algorithmic}
\caption{Bayesian tensor training procedure}
\label{alg:bayes_EM_train}
\end{algorithm}
\subsubsection{Calibration metric}

We evaluate the overal calibration of our variational model using the sum 

\begin{equation}
    \Xi = \sum_{\alpha\in \Omega}|\xi_{1-2\alpha}-(1-2\alpha)|
\end{equation} 
of the \textit{calibration rate} $\xi_{1-2\alpha}$, which we define as

\begin{equation}
    \xi_{1-2\alpha} = \frac{\text{number of $y_{nmt}$ within $1-2\alpha$ confidence level}}{\text{total number of $y_{nmt}$}}
\end{equation}
where we consider $\alpha$ to take values in $\{0.05,0.15,0.25,0.35,0.45\}$.
Ideally, our model would provide a posterior predictive such that the calibration rate is

\begin{equation}
\xi_{1-2\alpha} = 1-2 \alpha
\end{equation}
for all $\alpha \in [0,1]$. To ensure that our model finds a meaningful variational approximation, we take our hyperparameter selection criteria to be:

\begin{equation}
    \eta_{\text{criteria}} = \Xi - R^2
    \label{cal_obj}
\end{equation}
where $R^2$ is calculated using $\Bigg( \prod_{p=1}^P \times_{-1} ( \Mb_p' \circ (\Mb_p\times_{2} \Kb^p))\Bigg)$ or $\Bigg(\prod_{p=1}^P (\Mb_p^s \circ (\Mb_p \times_2 \Kb^p) + \Mb_p^b) \Bigg)$. If we only use $\eta_{\text{criteria}}=\Xi$, we argue that there is an inductive bias in choosing $\alpha$'s which may lead to an approximation that is calibrated per se, but not meaningful as the modes are incorrect.
\section{Experimental setup and justification}
We divide our experiments into two parts, the first comparing our proposed frequentist model to LightGBM and FFM in terms of predictive performance on three datasets with high dimensional covariates. The second part is analyzing our proposed Bayesian model and its performance in  terms of calibration and predictive performance.  

\subsection{Frequentist experiments}
Our goal here is to showcase the performance of our proposed model in comparison to the more established LightGBM and FFM. We define our proposed model to be successful if it can soundly beat FFM and achieve on-par performance with LightGBM. In particular, LightGBM is a challenging benchmark as it has continuously received development, engineering and performance optimization since its inception in 2017. We execute our experiments by running 20 hyperopt \cite{Bergstra_hyperopt:a} iterations for all methods to find the optimal hyperparameter configuration constrained with a computational budget of 1000 iterations/gradient updates and a memory budget of approximately 16gb (which is the memory limit of a high-end GPU) for 5 different seeds, where the seed controls the randomization of how the data is split. We split our data into 60\% training, 20\% validation and 20 \% testing. We measure the performance of all models in terms of test data $R^2$. We compare the performance over the datasets Retail Fashion ($\sim 42$ million observations), Movielens-20M ($\sim 12$ million observations) \cite{10.1145/2827872} and Alcohol Sales ($\sim 3$ million observations). For details on hyperparameter choices, data and data preparation, please refer to the appendix. 

\begin{table}[t]
\begin{tabular}{llll}
\hline
Method/Dataset& Retail Sales & Movielens  & Alcohol Sales \\ \hline
$P$-way, WLR, Primal & 0.108$\pm$0.435& 0.38$\pm$0.012& 0.69$\pm$0.015\\
$P$-way, WLR, Dual&  0.571$\pm$0.007            &\bf{0.384$\pm$0.002}\footnotemark &\bf{0.704$\pm$0.013} \\
2-way, time only , WLR, Primal & 0.466$\pm$0.034&0.168$\pm$0.013  & 0.472$\pm$0.025 \\
2-way, time only , WLR, Dual &\bf{0.594$\pm$0.005}&0.349$\pm$0.01& 0.692$\pm$0.013\\ 
$P$-way, LS, Primal &0.152$\pm$0.033 &0.272$\pm$0.032  &0.542$\pm$0.026\\
$P$-way, LS, Dual &0.027$\pm$0.006& 0.281$\pm$0.027 &0.562$\pm$0.022\\ 
2-way, time only, LS, Primal &0.505$\pm$0.008&0.065$\pm$0.042&0.466$\pm$0.022\\
2-way, time only,LS, Dual &0.458$\pm$0.015&0.326$\pm$0.024 &0.699$\pm$0.012\\ \hline

\multicolumn{4}{c}{Benchmark models}                                \\ \hline
LightGBM                  & 0.581$\pm$0.009   & 0.303$\pm$0.003 & 0.646$\pm$0.015   \\ 
FFM&0.48$\pm$0.032& 0.335$\pm$0.055            & 0.64$\pm$0.019\\ 
Linear &0.058$\pm$0.001 &0.185$\pm$0 &  0.03$\pm$0.002 \\ \hline
\end{tabular}
\caption{Frequentist experiment results. Performance measured in $R^2$}
\label{table:freq_exp}
\end{table}
\footnotetext{RMSE=0.825$\pm$0.002, which is comparable to state-of-the-art neural methods presented in \cite{10.5555/3045390.3045472}}
We see in table \ref{table:freq_exp} that KFT has configurations that can outperform the benchmarks with a good margin and that the standard deviations are not overlapping. 

\subsection{Bayesian experiments}

We use the same setup as in the frequentist case, modulo the hyperparameter evaluation objective which instead is \eqref{cal_obj}. For details on hyperparameter choices and data preparation, please refer to the appendix. In table \ref{table:Bayesian_results} we find that the we are able to obtain fairly calibrated variational approximations with the retail dataset having the best performance. By using a Bayesian framework we seem to generally lose some predictive performance except in the case of Movielens.  

\begin{figure}[]
\centerline{\includegraphics[width=0.8\columnwidth]{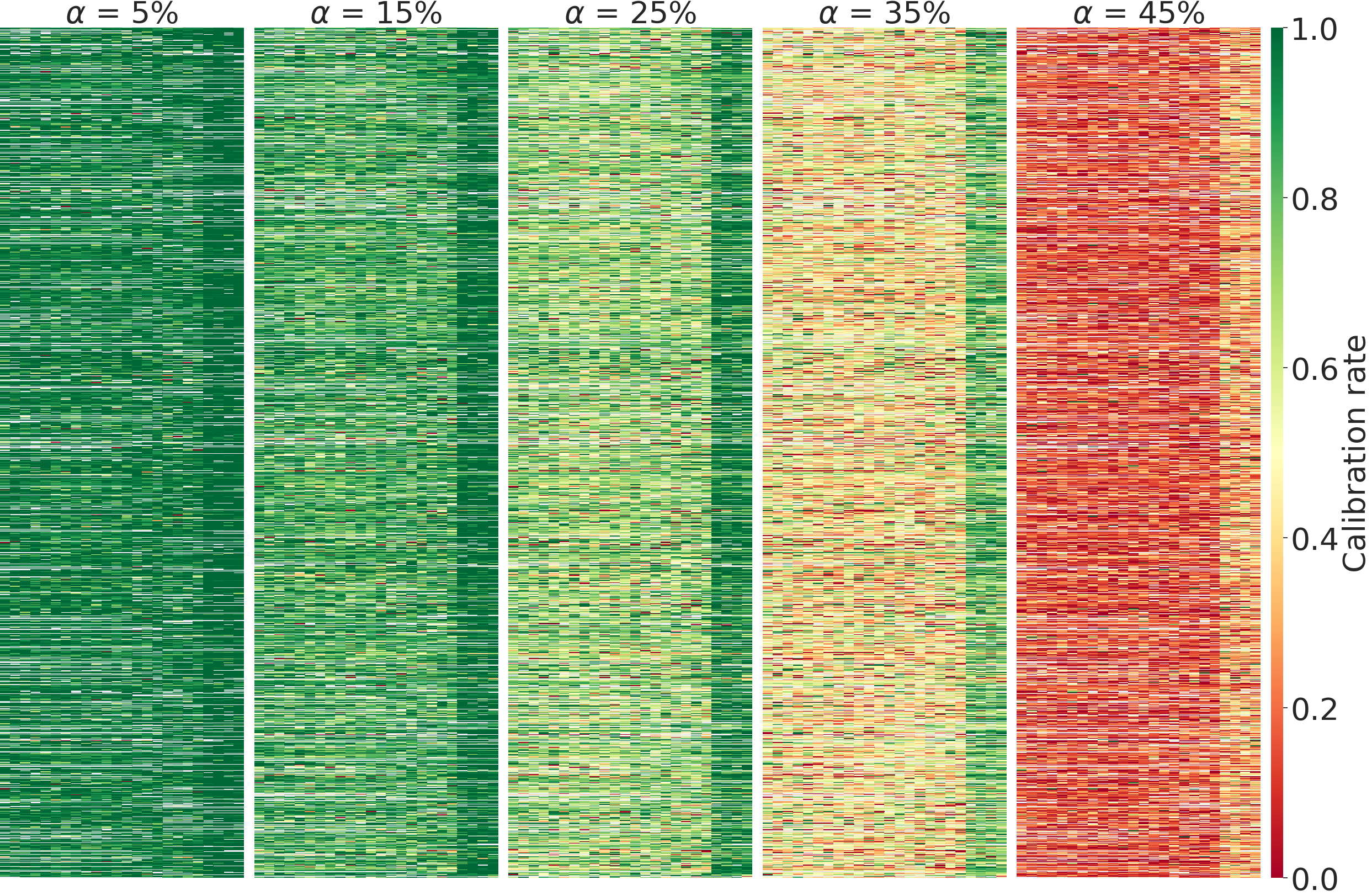}}
\caption{Heatmat of calibration rates for Alcohol sales. Here the y-axis is location and x-axis is time, where sales have been aggregated on items. We see that the calibration rate over all aggregates consistently adjusts with changes in $\alpha$}
\label{alcohol_heatmap}
\end{figure}

\begin{figure}[]
\centerline{\includegraphics[width=0.8\columnwidth]{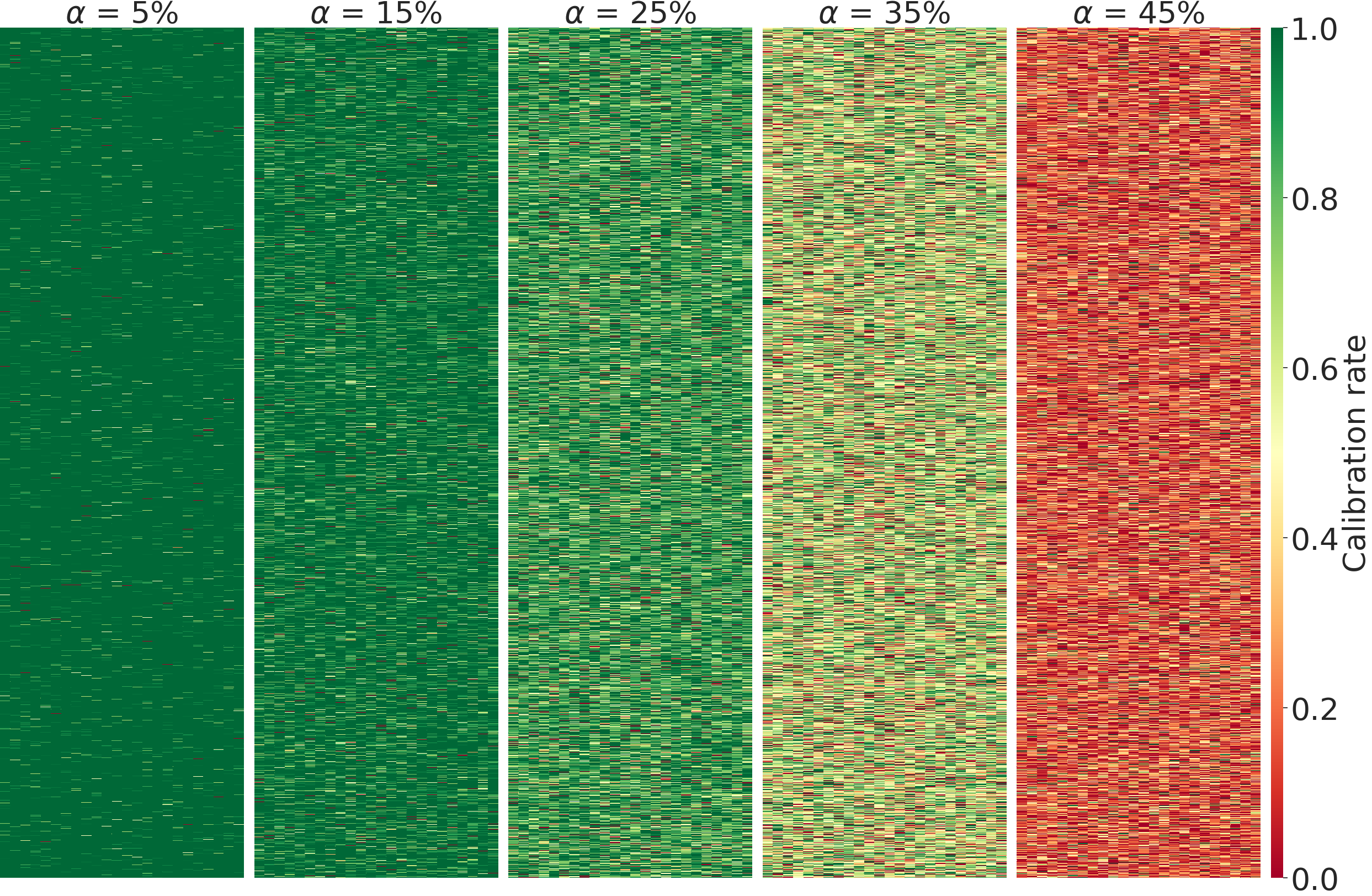}}
\caption{Heatmat of calibration rates for Movielens. Here the y-axis is users and x-axis is time, where ratings have been aggregated on movies. We see that the calibration rate over all aggregates consistently adjusts with changes in $\alpha$}
\label{movielens_heatmap}
\end{figure}

\begin{figure}[]
\centerline{\includegraphics[width=0.8\columnwidth]{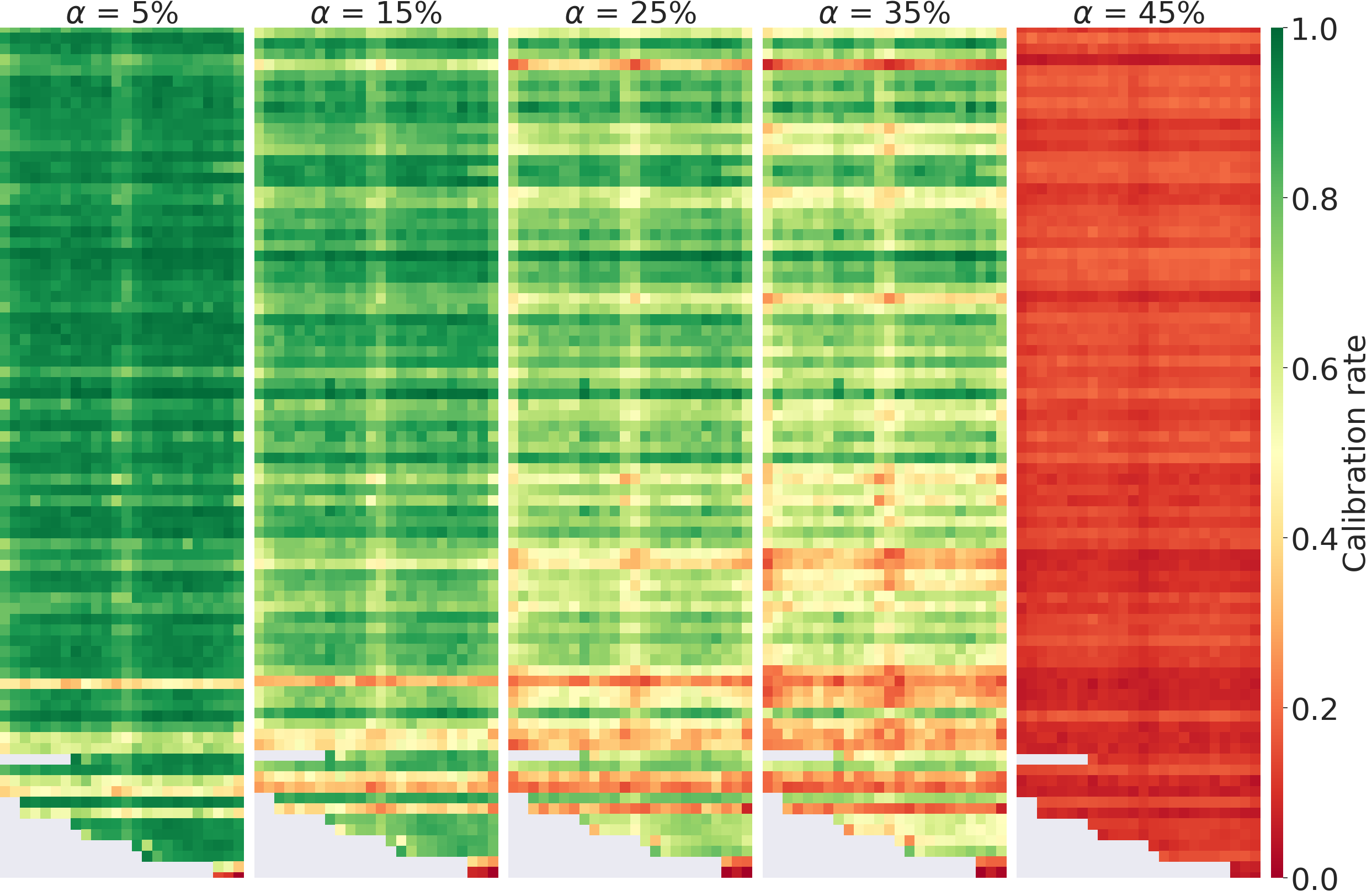}}
\caption{Heatmat of calibration rates for Retail sales. Here the y-axis is store and x-axis is time, where sales have been aggregated on articles. We see that the calibration rate over all aggregates consistently adjusts with changes in $\alpha$}
\label{retail_heatmap}
\end{figure}

\begin{table}[]
\resizebox{\columnwidth}{!}{%
\begin{tabular}{lccccccccr}
\midrule
  Data set&Model&$\Xi_{0.05}$ & $\Xi_{0.15}$ & $\Xi_{0.25}$ &$\Xi_{0.35}$  & $\Xi_{0.45}$ & $\Xi$&$R^2$&$\eta_{\text{criteria}}$\\
 \midrule
  \makecell{Alcohol \\Sales}&\makecell{\textbf{$P$-way, WLR, Dual}\\\textbf{Univariate} }& \makecell{0.056\\$\pm$0.012}& \makecell{ 0.181\\$\pm$0.028} &\makecell{0.256\\$\pm$0.044} & \makecell{ 0.249\\$\pm$0.050}  &\makecell{ 0.113\\$\pm$ 0.027} &\makecell{0.858\\$\pm$0.160} &\makecell{ 0.624\\$\pm$0.024}&\makecell{0.231 \\$\pm$0.149}\\
   \makecell{Alcohol \\Sales}&\makecell{$P$-way, WLR, Dual\\Multivariate}& \makecell{0.087\\$\pm$0.008}& \makecell{0.259 \\$\pm$0.024} &\makecell{0.397\\$\pm$0.051} & \makecell{ 0.442\\$\pm$0.089}  &\makecell{ 0.227\\$\pm$ 0.065} &\makecell{ 1.412 \\$\pm$0.236  } &\makecell{0.606\\$\pm$0.023}&\makecell{0.806\\$\pm$0.231}\\
     \makecell{Alcohol \\Sales}&\makecell{$P$-way, LS, Dual\\Univariate }& \makecell{0.097\\$\pm$0.004}& \makecell{0.290 \\$\pm$0.011} &\makecell{0.475\\$\pm$0.023} & \makecell{ 0.629 \\$\pm$0.052}  &\makecell{ 0.461\\$\pm$ 0.072} &\makecell{1.952\\$\pm$ 0.160   } &\makecell{0.458\\$\pm$0.034}&\makecell{1.494\\$\pm$0.149}\\
     \makecell{Alcohol \\Sales}&\makecell{$P$-way, LS, Dual\\Multivariate }& \makecell{0.098\\$\pm$0.004}& \makecell{0.294 \\$\pm$0.011} &\makecell{ 0.487\\$\pm$0.026 } & \makecell{ 0.659 \\$\pm$0.069}  &\makecell{ 0.508\\$\pm$0.119} &\makecell{2.046\\$\pm$0.228   } &\makecell{-5.398\\$\pm$7.163}&\makecell{7.444\\$\pm$6.946 }\\
           \makecell{Movielens}&\makecell{$P$-way, WLR, Dual\\Univariate }& \makecell{0.1\\$\pm$0.0}& \makecell{0.299 \\$\pm$0.001 } &\makecell{0.478\\$\pm$0.019} & \makecell{ 0.507\\$\pm$0.086}  &\makecell{  0.186 \\$\pm$  0.047} &\makecell{ 1.569 \\$\pm$  0.152    } &\makecell{0.293\\$\pm$0.022}&\makecell{ 1.276\\$\pm$0.167}\\
      \makecell{Movielens}&\makecell{\textbf{$P$-way, WLR, Dual}\\\textbf{Multivariate} }& \makecell{0.097\\$\pm$0.003}& \makecell{0.274 \\$\pm$0.016} &\makecell{0.390\\$\pm$0.04} & \makecell{ 0.360\\$\pm$0.060}  &\makecell{ 0.145\\$\pm$ 0.033} &\makecell{ 1.266 \\$\pm$ 0.148  } &\makecell{0.387\footnotemark \\$\pm$ 0.003}&\makecell{0.879\\$\pm$0.145}\\
     \makecell{Movielens}&\makecell{$P$-way, LS, Dual\\Univariate }& \makecell{0.1\\$\pm$0.0}& \makecell{0.299 \\$\pm$0.001 } &\makecell{0.466\\$\pm$0.021} & \makecell{0.440\\$\pm$0.099}  &\makecell{0.163  \\$\pm$0.084} &\makecell{  1.468 \\$\pm$  0.203     } &\makecell{ 0.308\\$\pm$0.010}&\makecell{  1.161 \\$\pm$0.206}\\
      \makecell{Movielens}&\makecell{$P$-way, LS, Dual\\Multivariate }& \makecell{0.1\\$\pm$0.0}& \makecell{0.300 \\$\pm$0.0 } &\makecell{0.491\\$\pm$0.005} & \makecell{0.576\\$\pm$0.015}  &\makecell{0.277  \\$\pm$0.025 } &\makecell{1.744\\$\pm$0.037 } &\makecell{ 0.323\\$\pm$0.001}&\makecell{  1.421 \\$\pm$0.038}\\
   \makecell{Retail\\Sales}&\makecell{2-way, WLR, Dual\\time only, Univariate}& \makecell{0.021\\$\pm$0.010}& \makecell{0.061 \\$\pm$0.046 } &\makecell{0.085\\$\pm$0.055} & \makecell{0.071\\$\pm$0.045}  &\makecell{0.027   \\$\pm$0.018 } &\makecell{0.265 \\$\pm$0.169 } &\makecell{ 0.506\\$\pm$0.061}&\makecell{  -0.241 \\$\pm$0.207}\\
      \makecell{Retail\\Sales}&\makecell{2-way, WLR, Dual\\time only, Multivariate }& \makecell{0.153\\$\pm$0.042}& \makecell{0.123\\$\pm$0.045 } &\makecell{0.080 \\$\pm$0.041} & \makecell{0.041\\$\pm$0.029}  &\makecell{ 0.013   \\$\pm$0.009 } &\makecell{0.411  \\$\pm$0.163  } &\makecell{ 0.510\\$\pm$0.040}&\makecell{  -0.099 \\$\pm$0.179}\\
    \makecell{Retail\\Sales}&\makecell{2-way, LS, Dual\\time only, Univariate  }& \makecell{ 0.077 \\$\pm$0.014}& \makecell{0.221\\$\pm$0.035 } &\makecell{0.310  \\$\pm$0.055} & \makecell{0.298\\$\pm$0.052 }  &\makecell{ 0.129 \\$\pm$0.020 } &\makecell{ 1.036  \\$\pm$0.169  } &\makecell{0.498\\$\pm$0.056}&\makecell{0.538\\$\pm$0.147}\\
    \makecell{Retail\\Sales}&\makecell{\bf{2-way, LS, Dual}\\\bf{time only, Multivariate}}& \makecell{ 0.058 \\$\pm$0.021}& \makecell{0.022\\$\pm$0.014 } &\makecell{0.041  \\$\pm$0.028} & \makecell{0.051 \\$\pm$0.020 }  &\makecell{ 0.023 \\$\pm$0.007 } &\makecell{ 0.195\\$\pm$0.031  } &\makecell{0.469\\$\pm$0.064}&\makecell{-0.274 \\$\pm$0.090}\\
 \midrule
\end{tabular}
}
\caption{Table of results for Bayesian experiments. Here we denote $\Xi_{\alpha} =|\xi_{1-2\alpha} - (1-2\alpha)|$. We ideally want $\Xi_{\alpha}$ and $\Xi$ to be as close to zero as possible coupled with a high $R^2$. }
\label{table:Bayesian_results}
\end{table}
\footnotetext{RMSE=0.821$\pm$0.0, which is lower than comparable Bayesian methods introduced in \cite{10.1145/1553374.1553452}\cite{Luo2016}}
\noindent We provide a visualization of the calibration ratio in figure \ref{alcohol_heatmap},\ref{movielens_heatmap} and \ref{retail_heatmap}.

\section{Analysis}

We demonstrated the practical utility of KFT in both a frequentist and Bayesian context. We now scrutinize the robustness and effectiveness of KFT as a remedy for constraint-imposing side information.

\subsection{Does adding side information really improve performance?}
To uncover this, we run a naive tensor model $P$-way factorization without any side information. The first row in table \ref{table:additional_exp} shows the results for all three datasets. We find that a naive model with no side information is unable to match the performance of a model with full side information for all three datasets, indicating that adding side information is indeed a meaningful endeavor.
\subsection{Does KFT actually alleviate side information imposed constraints?}
We run a $P$-way model with kernelized side information over 5 seeds with results in the second row of table \ref{table:additional_exp}. We see that the experimental results indeed align with the proposed theoretical constraint of adding side information as the performance across all datasets are poor, which empirically demonstrates the need for KFT.
\subsection{How does KFT perform when applying constant side information?}
To answer this question, we replace all side information with a constant $\mathbf{1}$ and kernelize it. The results in the third row of table \ref{table:additional_exp} indicate that KFT indeed is robust towards constant side information, as the performance does not degrade dramatically. 
\subsection{How does KFT perform when applying noise as side information?}
Similar to the previous question, we now replace the side information with standard Gaussian noise instead. The results in the last row of table \ref{table:additional_exp} indicate that KFT also is robust against noise and surprisingly performant as well. A possible explanation for this is that adding Gaussian noise serves as an implicit regularizer or that the original side information is similarly distributed as standard Gaussian noise. We conclude that KFT is stable against uninformative side information in the form of Gaussian noise. 

\begin{table}[t]
\begin{tabular}{llll}
\hline
Method/Dataset& Retail Sales & Movielens  & Alcohol Sales \\ \hline
Vanilla Tensor, No side information& 0.385$\pm$0.002&0.147$\pm$0.038 &0.477$\pm$0.015 \\
Vanilla Tensor, Side information &-0.087$\pm$0.017&-4.837$\pm$0.028& 0.196$\pm$0.008\\ 
WLR, Constant side information, Dual &0.56$\pm0.013$& 0.33$\pm$0.039& 0.605$\pm$0.041\\ 
WLR, Noise side information, Dual &0.553$\pm$0.006& 0.388$\pm$0.003 & 0.704$\pm$0.01\\ \hline
\end{tabular}
\caption{Table with results from additional experiments for $P$-way dual models}
\label{table:additional_exp}
\end{table}

\section{Conclusion}

We identified an inherent limitation of side information based tensor regression and gave a method which removes this limitation. Our proposed KFT method yields competitive performance against state-of-the-art industrial forecasting models on a fixed computational budget. Specifically, as the experiments in table~\ref{table:freq_exp} demonstrate, for at least some cases of real practical interest, WLR is the most performant configuration. Further, KFT offers extended versatility in terms of calibrated Bayesian variational estimates. Our analysis shows that KFT solves the problems we described in section~\ref{section:backg} and is robust for adversarial side information in the form of Gaussian noise. A direction for further development would be to characterise identifiability conditions for KFT and extend the Bayesian framework beyond mean-field variational inference.

\clearpage
\nocite{langley00}
\bibliography{example_paper}
\bibliographystyle{spmpsci} 

\clearpage

\appendix
\section{Data and data processing}

\subsection{Data Processing}
Our philosophy in this paper is to limit data processing to simple operations that does not require excessive engineering for a fair comparison in both utility and input data. For all the models, we carry out the exact same preprocessing modulo the models requirements of data format. We do the following general processing steps: 
\begin{enumerate}
    \item Extract relevant features and parse them to be continuous or categorical.
    \item Scale all features using a z-transformation
\end{enumerate}
For each model specifically we do the following:
\begin{enumerate}
    \item KFT: Tensorize all data by expressing all main modes (i.e. person, movie, time etc) as a tensor with side information associated with each mode.
    \item LightGBM: Here, we don't scale the features as boosting trees generally performs better with unscaled data. In some cases we have applied PCA to some of the side information that was joined on the data matrix to decrease the memory footprint of the data matrix to contain it to a practical size. 
    \item FFM: Here we bin all continuous features, as FFM requires all data to be categorical. 
    \item Linear regression: For large categorical features, we using feature hashing to avoid data matrices of infeasible sizes. All other categorical features get one-hot encoded.
\end{enumerate}

\subsection{Retail Sales Data}
\begin{table}[]
\caption{Dataset description for retail sales data. Note that we only present the ID of each feature to preserve anonymity and that the examples are not real.}
\label{retail_data_features}
\label{data_retail}
\vskip 0.15in
\begin{small}
\begin{sc}
\begin{tabular}{lcccr}
\toprule
property              & unit & quantity \\
\midrule
monthly sales    & count   & 42490633 \\
unique location\_id's & count   & 80      \\
unique article\_id's  & count   & 463763   \\
\midrule 
article\_id side info    & description   &example \\
\midrule
appearance\_id &article style& 26    \\ 
colour\_id  &article color& 1337   \\
product\_id & article group& 15\\
size\_id & article size & L\\
department\_id &article group theme &teen\\
product\_season\_id&article group season & spring\\
product\_type\_id&product group type& shirt\\
product\_group\_no&product group no& accessories\\
price & article price & 5usd\\
\midrule 
location\_id side info    & description   &example \\
\midrule
city\_id  &city of store& 56   \\
longitude  &longitude& 34.23   \\
latitude  &latitude& 34.24   \\
brand\_id & brand & 12\\
opening\_year\_id & opening year&123\\
opening\_month\_id &opening month &92 \\
opening\_day\_id &opening day &97 \\
opening\_hours\_mo-su &hours open mo-su & 12,..,10\\
no\_of\_floors &store no. of floors & 10\\
total\_sales\_area & store area size & 1231sqm\\
special\_sales\_area & special sales area & 789sqm\\
\midrule 
time side info    & description   &example \\
\midrule
year  &year& 2019   \\
month  &month&1   \\
\bottomrule
\end{tabular}
\end{sc}
\end{small}
\vskip -0.1in
\end{table}

We detail the features of the Retail Sales data in table \ref{retail_data_features}. Here we choose our modes to be store, articles and time.  
\subsection{Movielens}

We detail the features of the Retail Sales data in \ref{movielens}. Here we choose our modes to be users, movies and time of rating given. For the Movielens data, it should be noted that we filter the movies on existing entries in the side information. This is why we only have roughly 11 million observations rather than 20 million. 

\begin{table}[]
\caption{Dataset description of Movielens}
\label{movielens}
\vskip 0.15in
\begin{small}
\begin{sc}
\begin{tabular}{lcccr}
\toprule
property              & unit & quantity \\
\midrule
ratings    & count   & 11880265 \\
users & count   & 138493      \\
movie\_id  & count   & 10370   \\
\midrule 
movie\_id side info   & description   &example \\
\midrule
genome\_score\_1  &genome score dimension 1& 0.345   \\
 \vdots & \vdots& \vdots \\
 genome\_score\_1000  &genome score dimension 1000& 0.1337   \\
\midrule 
time side info    & description   &example \\
\midrule
hour  &hour ratings was given & 12   \\
\bottomrule
\end{tabular}
\end{sc}
\end{small}
\vskip -0.1in
\end{table}

\subsection{Alcohol Sales}

\begin{table}[]
\caption{Dataset description of Iowa alcohol sales}
\label{data_public}
\vskip 0.15in
\begin{small}
\begin{sc}
\begin{tabular}{lcccr}
\toprule
property              & unit & quantity \\
\midrule
bottles sold    & count   & 3036063 \\
unique store\_location\_id's & count   & 3476      \\
unique item\_id's  & count   & 4542   \\
\midrule 
item\_id side info    & description   &example \\
\midrule
category &type of alcohol& irish whiskey    \\ 
pack  &size of package& 6   \\
bottle volume (ml) & bottle volume& 750\\
state bottle cost& cost for retailer to buy &4usd\\
state bottle retail  &retail price &5usd\\
\midrule 
store\_location\_id side info    & description   &example \\
\midrule
city\_id  &city of store& 56   \\
longitude  &longitude& 34.23   \\
latitude  &latitude& 34.24   \\
county & county &shelby\\
store\_number & store number &12\\
zip code &zip code & 157\\
\midrule 
time side info    & description   &example \\
\midrule
year  &year& 2019   \\
month  &month&1   \\
\bottomrule
\end{tabular}
\end{sc}
\end{small}
\vskip -0.1in
\end{table}
We detail the features of the Alcohol Sales data in \ref{data_public}. Here we choose our modes to be location, item and time.  

\section{Hyperparameter configuration}

\subsection{Frequentist hyperparameters}
We run all frequentist experiments for 10 epochs with 100 iterations for each epoch with 20 hyperparameter search iterations. We consider two decomposition types:

\begin{enumerate}
    \item $P$-way latent factorization, where each dimension has a latent component. In principle, this can be thought of as each dimension being independently factorized. Further we utilize all possible side information
    \item 2-way latent factorization, where time is grouped with another dimension as one latent component and with the other dimensions grouped in a second latent component. Here we only consider time as side information, for the purpose of only modelling temporal changes. 
\end{enumerate}
Our models are generally searched over the configurations described in table \ref{KFT_hyperpara_frequentist}.
\begin{table}[H]
\begin{tabular}{l|l}
\hline
\textbf{Property} & \textbf{Range/Choices}                      \\ \hline
batch size\footnotemark        & {[}0.01,0.1{]}                              \\
learning rate     & \{1e-3,1e-2,1e-1\}                          \\
R\footnotemark& {[}5,70{]}                                  \\
$\lambda_p$       & {[}0,1{]}                                   \\
$\lambda_p'$      & {[}0,1{]}                                   \\
kernel choice     & \{rbf, matern 0.5, matern 1.5, matern 2.5\} \\ \hline
\end{tabular}
\caption{}
\label{KFT_hyperpara_frequentist}
\end{table}
\noindent For exact details we refer to the \href{https://github.com/MrHuff/KernelFriedTensor/}{code base}.
\footnotetext{The upper bound varies for each dataset, as memory limitation changes with dataset}
\footnotetext{Batch size expressed as a proportion of total training data examples. As an example our smallest batch size use 1\% of the training data as the batch size.}

\subsubsection{LightGBM hyperparameters}
We provide hyperparameters for LightGBM in table \ref{LightGBMhyper}

\begin{table}[H]
\begin{tabular}{l|l}
\hline
\textbf{Property}       & \textbf{Range/Choices}      \\ \hline
num leaves              & {[}7,4095{]}                \\
learning rate           & {[}$\exp(-5),\exp(-2.3)${]} \\
min data in leaf        & {[}10,30{]}                 \\
min sum hessian in leaf & {[}$\exp(0),\exp(2.3)${]}   \\
bagging freq            & {[}1,5{]}                   \\
$\lambda_1$              & {[}0,10{]}                  \\
$\lambda_2$              & {[}0,10{]}                  \\ \hline
\end{tabular}
\caption{}

\label{LightGBMhyper}
\end{table}

\subsubsection{FFM hyperparameters}
We provide hyperparameters for FFG in table \ref{FFM}

\begin{table}[H]
\begin{tabular}{l|l}
\hline
\textbf{Property} & \textbf{Range/Choices} \\ \hline
R                 & {[}2,20{]}             \\
batch size        & {[}1e-6,0.1{]}         \\
learning rate     & {[}0.05,1.0{]}         \\
$\lambda$         & {[}0,0.005{]}          \\ \hline
\end{tabular}
\caption{}

\label{FFM}
\end{table}

\subsubsection{Linear regresison hyperparameters}
We provide hyperparameters for FFG in table \ref{linear}

\begin{table}[H]
\begin{tabular}{l|l}
\hline
\textbf{Property} & \textbf{Range/Choices} \\ \hline
batch size        & {[}1e-6,0.1{]}         \\
learning rate     & {[}0.05,1.0{]}         \\
$\lambda$         & {[}0,1.0{]}          \\ \hline
\end{tabular}
\caption{}

\label{linear}
\end{table}

\subsection{Bayesian hyperparameters}
We run all frequentist experiments for 20 epochs with 100 iterations for each epoch with 10 hyperparameter search iterations. We list the hyperparameters in table \ref{KFT_hyperpara_Bayesian}

\begin{table}[H]
\begin{tabular}{l|l}
\hline
\textbf{Property} & \textbf{Range/Choices}                      \\ \hline
batch size        & {[}0.01,0.05{]}                              \\
learning rate     & \{1e-3,1e-2,1e-1\}                          \\
R& [5,30]                                  \\
$\sigma_Y$       & {[}1e-3,100{]}                                   \\
$\mu_p$       & {[}0,0{]}                                   \\
$\mu_p'$       & {[}0,0{]}                                   \\
$\sigma_p^2$       & {[}$\exp(-1),\exp(3)${]}                                   \\
$\sigma_p'$       & {[}$\exp(-1),\exp(3)${]}                                   \\
kernel choice     & \{rbf, matern 0.5, matern 1.5, matern 2.5\} \\ \hline
\end{tabular}
\caption{}
\label{KFT_hyperpara_Bayesian}
\end{table}
\noindent For exact details we refer to the \href{https://github.com/MrHuff/KernelFriedTensor/}{code base}.

\section{Derivation for weighted latent regression regularization term}
\begin{equation}
    \begin{split}
        &\sum_{r_1r_2}\Vert \vsf_r\Vert^{2}_{\mathcal{H}}=\sum_{r_1r_2} \langle\vsf_r,\vsf_r \rangle_{\mathcal{H}}\\&=\sum_{r_1r_2}\sum_{\substack{n_1,...,n_q \\n_1',...,n_q'}}v_{r_1n_1...n_qr_1}v_{r_1n_1'...n_q'r_2}'\prod_{q=1}^{Q}\delta_{\xb_{n_q'}^{Q}}(\cdot)\prod_{q=1}^{Q}k(\cdot,\xb_{n_q}^{Q})  \\ &\sum_{\substack{\bar{n}_,...,\bar{n}_q \\\bar{n}_1',...,\bar{n}_q'}}v_{r_1\bar{n}_1...\bar{n}_qr_2}v_{r_1\bar{n}_1'...\bar{n}_q'r_2}'\prod_{q=1}^{Q}\delta_{\xb_{\bar{n}_q'}^{Q}}(\cdot)\prod_{q=1}^{Q}k(\cdot,\xb_{\bar{n}_q}^{Q}) \\=&\sum_{r_1r_2}\sum_{\substack{n_1,...,n_q\\n_1',...,n_q'\\\bar{n}_1,...,\bar{n}_q}}{v'_{r_1n_1',...,n_q' r_2}}^2v_{r_1 n_1...n_qr_2}v_{r_1\bar{n}_1...\bar{n}_qr_2}\prod_{q=1}^Q k(\xb_{n_q}^{Q},\xb_{\bar{n}_q}^{Q})\\=&\text{tr}\Bigg((\prod_{q=1}^Q\otimes\Kb^{Q}) \cdot \Vb_{R_1\cdot R_2\times{\prod_{q=1}^Qn_q}}^{\top}\cdot\left( (\prod_{q=1}^Q(\Vb'\circ\Vb')\times_q \mathbf{1}_{n_q\times n_q})\circ \Vb  \right)_{R_1\cdot R_2\times{\prod_{q=1}^Qn_q}} \Bigg) \\ = & \text{tr}\Bigg((\prod_{q=1}^Q \Vb\times_{q+1}\Kb^q)_{R_1\cdot R_2\times{\prod_{q=1}^Qn_q}}^{\top}\cdot  \left( (\prod_{q=1}^Q(\Vb'\circ\Vb')\times_q \mathbf{1}_{n_q\times n_q})\circ \Vb  \right)_{R_1\cdot R_2\times{\prod_{q=1}^Qn_q}} \Bigg) \\ =&\Bigg((\prod_{q=1}^Q \Vb\times_{q+1}\Kb^q) \circ \left( (\prod_{q=1}^Q(\Vb'\circ\Vb')\times_q \mathbf{1}_{n_q\times n_q})\circ \Vb  \right)\Bigg)_{++}
    \end{split}
\end{equation}

\subsection{RFF regularization term}

The regularization term can similarly be generalized to

\begin{equation}
\begin{split}
        \sum_{r_1r_2}\Vert \vsf_r\Vert^{2}_{\mathcal{H}}&=\sum_{r_1r_2}\sum_{\substack{n_1,...,n_q\\n_1',...,n_q'\\\bar{n}_1,...,\bar{n}_q}}{v'_{r_1n_1',...,n_q'r_2}}^2v_{r_1n_1...n_qr_2}v_{r_1\bar{n}_1...\bar{n}_qr_2}\prod_{q=1}^Q k(\xb_{n_q}^{Q},\xb_{\bar{n}_q}^{Q})\\&\approx\sum_{r_1r_2}\sum_{\substack{n_1,...,n_q\\n_1',...,n_q'\\\bar{n}_1,...,\bar{n}_q}}{v'_{r_1n_1',...,n_q'r_2}}^2\prod_{q=1}^Q (v_{r_1n_1...n_qr_2}\phi(\xb_{n_q}^Q))^{\top}( v_{r_1\bar{n}_1...\bar{n}_qr_2}\phi(\xb_{\bar{n}_q}^Q))\\ &=\Bigg((\prod_{q=1}^Q \Vb\times_{q+1} \Phi_{q}\times_q \Phi_q^{\top} ) \circ \left( (\prod_{q=1}^Q(\Vb'\circ\Vb')\times_q \mathbf{1}_{n_q\times n_q})\circ \Vb  \right)\Bigg)_{++}
\end{split}
\end{equation}

\section{Training procedure motivation}

Consider a data matrix $\frac{\Xb}{\sigma}\in \RR^{N\times d}$ where $\sigma$ is a hyperparameter that controls the scaling of $\Xb$ and a target $\Yb \in \RR^{N}$. Assume $N$ is too large and we have to resort to first order stochastic gradient methods to approximate $\mu\in \RR^d$ in our regression $\frac{\Xb}{\sigma})\mu\approx \Yb$. Exploiting autograd \cite{paszke2017automatic}, together with ADAM \cite{kingma2014adam} we can in practice optimize $\{\sigma,\mu\}$ simultaneously for each iteration. However doing this, we commit a fallacy as when updating $\mu_t = \mu_{t-1}-\frac{\partial \mathcal{L}(\sigma_{t-1},\mu_{t-1})}{\partial \mu_{t-1}}$ and $\sigma_t = \sigma_{t-1}-\frac{\partial \mathcal{L}(\sigma_{t-1},\mu_{t-1})}{\partial \sigma_{t-1}}$, we do not account for the mixed partial $\frac{\partial}{\partial \sigma_{t-1}}\frac{\partial \mathcal{L}(\sigma_{t-1},\mu_{t-1})}{\partial \mu_{t-1}} \propto \frac{-1}{\sigma_{t-1}^3}$ assuming mean square error. Thus updating $\{\sigma,\mu\}$ simultaneously using first order derivatives would yield an update error for $\mu$ as the updating gradient does not adjust for the mixed partial  $\frac{\partial}{\partial \sigma_{t-1}}\frac{\partial \mathcal{L}(\sigma_{t-1},\mu_{t-1})}{\partial \mu_{t-1}}$ when $\sigma$ is being updated at the same time. This scenario extends almost one-to-one for KFT, as we would commit a similar fallacy by updating all parameters at once. Hence we take an EM-approach when updating $\l_p,\Vb_p,\Vb_p'$. 
\section{$\mathbb{E}_{q}[\log{p(y_{i_1,...,i_P}|\Vb_1,...,\Vb_p)}]$ derivations}

\subsection{Regression}

\subsubsection{Weighted Latent regression}
Assuming that $\sigma_y$ is a constant hyperparameter. We first have that 
\begin{equation}
\begin{split}
    &\mathbb{E}_{q}[\log{p(y_{i_1,...,i_P}|\Vb_1,...,\Vb_p,\Vb_1',...,\Vb_p')}] \\ \propto & \frac{1}{\sigma_y^2} \mathbb{E}_{q}[(y_{i_1,...,i_P}-  \sum_{\substack{r_1,...,r_P\\i_1',...,i_P'\\i_1'',...,i_P''}} \prod_{p=1}^P v_{r_{p}i_p''r_{p-1}}'v_{r_{p}i_p'r_{p-1}}\underbrace{k(\xb_{i_p}^p,\xb_{i_p'}^p)\delta_{\xb_{i_p}^p}(\xb_{i_p''}^p)}_{f(k,\delta)} )^2 ] \\ &= \frac{1}{\sigma_y^2} \Bigg(y_{i_1,...,i_P}^2 - 2y_{i_1,...,i_P}\cdot  \sum_{\substack{r_1,...,r_P\\i_1',...,i_P'\\i_1'',...,i_P''}} \prod_{p=1}^P \mathbb{E}_{q}[v_{r_{p}i_p''r_{p-1}}']\mathbb{E}_{q}[v_{r_{p}i_p'r_{p-1}}]f(k,\delta) \\&+\sum_{\substack{r_1,...,r_P\\i_1',...,i_P'\\i_1'',...,i_P''\\q_1,...,q_P\\j_1',...,j_P'\\j_1'',...,j_P''}} \prod_{p=1}^P \mathbb{E}_{q}[v_{q_{p}j_p''q_{p-1}}' v_{r_{p}i_p''r_{p-1}}']\mathbb{E}_{q}[v_{r_{p}i_p'r_{p-1}}v_{q_{p}j_p'q_{p-1}}]\\&k(\xb_{i_p}^p,\xb_{i_p'}^p)\delta_{\xb_{i_p}^p}(\xb_{i_p''}^p)k(\xb_{i_p}^p,\xb_{j_p'}^p)\delta_{\xb_{i_p}^p}(\xb_{j_p''}^p)\Bigg)
\end{split}
\end{equation}
Our goal is to find the corresponding tensor operations of the sum terms.

\paragraph{Univariate meanfield}
We first have that

\begin{equation}
\begin{split}
        &\mathbb{E}_{q}[v_{r_{p}i_p'r_{p-1}}v_{q_{p}j_p'q_{p-1}}] =\mu_{r_{p}i_p'r_{p-1}}\mu_{q_{p}j_p'q_{p-1}} + \delta_{r_{p}}(q_{p})\delta_{r_{p-1}}(q_{p-1}) \delta_{i_{p}'}(j_p')\sigma_{r_{p}i_p'r_{p-1}}^2  
\end{split}
\end{equation}
Then the last term is can be written as
\begin{equation}
    \begin{split}
        &\Longrightarrow\sum_{\substack{r_1,...,r_P\\i_1',...,i_P'\\q_1,...,q_P\\j_1',...,j_P'}} \prod_{p=1}^P \mathbb{E}_{q}[v_{q_{p}i_pq_{p-1}}' v_{r_{p}i_pr_{p-1}}']\mathbb{E}_{q}[v_{r_{p}i_p'r_{p-1}}v_{q_{p}j_p'q_{p-1}}]k(\xb_{i_p}^p,\xb_{i_p'}^p)k(\xb_{i_p}^p,\xb_{j_p'}^p)\\= &\sum \prod_{p=1}^P (\mu_{q_{p}i_pq_{p-1}}' \mu_{r_{p}i_pr_{p-1}}' + \delta_{r_{p}}(q_{p})\delta_{r_{p-1}}(q_{p-1}) \sigma_{r_{p}i_pr_{p-1}}'^2 )(\mu_{q_{p}j_p'q_{p-1}} \mu_{r_{p}i_p'r_{p-1}} + \\& \delta_{r_{p}}(q_{p})\delta_{r_{p-1}}(q_{p-1}) \delta_{i_{p}'}(j_p')\sigma_{r_{p}i_p'r_{p-1}}^2 )k(\xb_{i_p}^p,\xb_{i_p'}^p)k(\xb_{i_p}^p,\xb_{j_p'}^p)\\&= \sum \prod_{p=1}^P\mu_{q_{p}i_pq_{p-1}}' \mu_{r_{p}i_pr_{p-1}}'\mu_{q_{p}j_p'q_{p-1}}\mu_{r_{p}i_p'r_{p-1}}k(\xb_{i_p}^p,\xb_{i_p'}^p)k(\xb_{i_p}^p,\xb_{j_p'}^p)\\&+\sum_{\substack{r_1,...,r_P\\i_1',...,i_P'}} \prod_{p=1}^P\sigma_{r_{p}i_pr_{p-1}}'^2\sigma_{r_{p}i_p'r_{p-1}}^2k(\xb_{i_p}^p,\xb_{i_p'}^p)^2\\&+\sum_{\substack{r_1,...,r_P\\i_1',...,i_P'\\j_1',...,j_P'}}\prod \sigma_{r_{p}i_pr_{p-1}}'^2\mu_{r_{p}i_p'r_{p-1}}\mu_{r_{p}j_p'r_{p-1}}k(\xb_{i_p}^p,\xb_{i_p'}^p)k(\xb_{i_p}^p,\xb_{j_p'}^p)\\&+\sum_{\substack{r_1,...,r_P\\i_1',...,i_P'}}\prod \sigma_{r_{p}i_p'r_{p-1}}^2\mu_{r_{p}i_pr_{p-1}}'^2k(\xb_{i_p}^p,\xb_{i_p'}^p)^2
    \end{split}
\end{equation}
\begin{equation}
    \begin{split}
                &\Longleftrightarrow \Bigg(\prod_p^P \times_{-1}   \Mb_p' \circ (\Mb_p\times_{2} \Kb^p)\Bigg)^2 + \Bigg(\prod_p^P \times_{-1}\Sigma_p' \circ (\Sigma_p\times_{2} (\Kb^p)^2)\Bigg)\\&+\Bigg(\prod_p^P \times_{-1}(\Sigma_p' \circ(\Mb_p\times_2 \Kb^p)^2) \Bigg)+\Bigg(\prod_p^P \times_{-1} \Mb_p'^2\circ (\Sigma_p\times_{2} (\Kb^p)^2)\Bigg)
    \end{split} 
\end{equation}
Where $\Mb_p',\Mb$ and $\Sigma',\Sigma$ correspond to the tensors containing the variational parameters $\mu_{r_{p}i_pr_{p-1}}'$, $\mu_{r_{p}i_pr_{p-1}}$ and $\sigma_{r_{p}i_pr_{p-1}}'^2$, $\sigma_{r_{p}i_pr_{p-1}}^2$ respectively. For the case of RFF's in dual space, we approximate $ \Sigma_p \times_2 (\Kb^p)^2 \approx \Sigma_p \times_2 (\Phi_p \bullet \Phi_p)^{\top} \times_2 (\Phi_p \bullet \Phi_p) $, where $\bullet$ is the transposed Khatri-Rao product. It should further be noted that any square term, means element wise squaring. With the middle term given by

\begin{equation}
    2\Yb \circ \Bigg( \prod_{p=1}^P \times_{-1} ( \Mb_p' \circ (\Mb_p\times_{2} \Kb^p))\Bigg)
\end{equation}
the full reconstruction term is expressed as:

\begin{equation}
    \begin{split}
        &\mathbb{E}_{q}[\log{p(y_{i_1,...,i_P}|\Vb_1,...,\Vb_p,\Vb_1',...,\Vb_p')}] \propto \\& \frac{1}{\sigma_y^2} \Bigg[\Yb^2-2\Yb \circ \Bigg( \prod_{p=1}^P \times_{-1} ( \Mb_p' \circ (\Mb_p\times_{2} \Kb^p))\Bigg) + \Bigg(\prod_p^P \times_{-1}   \Mb_p' \circ (\Mb_p\times_{2} \Kb^p)\Bigg)^2 \\&+ \Bigg(\prod_p^P \times_{-1}\Sigma_p' \circ (\Sigma_p\times_{2} (\Kb^p)^2)\Bigg)+\Bigg(\prod_p^P \times_{-1}(\Sigma_p' \circ((\Mb_p\times_2 \Kb^p)^2) \Bigg)\\&+\Bigg(\prod_p^P \times_{-1} \Mb_p'^2\circ (\Sigma_p\times_{2} (\Kb^p)^2)\Bigg) \Bigg]
    \end{split}
\end{equation}

\paragraph{Multivariate meanfield}
Similarly first observe that
\begin{equation}
\begin{split}
        &\mathbb{E}_{q}[v_{r_{p}i_p'r_{p-1}}v_{q_{p}j_p'q_{p-1}}] =\mu_{r_{p}i_p'r_{p-1}}\mu_{q_{p}j_p'q_{p-1}} +\delta_{r_{p}}(q_{p})\delta_{r_{p-1}}(q_{p-1}) \sigma_{i_p'}\sigma_{j_p'}.
\end{split}
\end{equation}
It then follows that the third term is calculated as
\begin{equation}
\begin{split}
&\sum \prod_{p=1}^P (\mu_{q_{p}i_pq_{p-1}}' \mu_{r_{p}i_pr_{p-1}}' + \delta_{r_{p}}(q_{p})\delta_{r_{p-1}}(q_{p-1}) \sigma_{r_{p}i_pr_{p-1}}'^2 )\\&(\mu_{q_{p}j_p'q_{p-1}} \mu_{r_{p}i_p'r_{p-1}} + \delta_{r_{p}}(q_{p})\delta_{r_{p-1}}(q_{p-1})\sigma_{i_p'}\sigma_{j_p'} )k(\xb_{i_p}^p,\xb_{i_p'}^p)k(\xb_{i_p}^p,\xb_{j_p'}^p)\\&= \sum \prod_{p=1}^P\mu_{q_{p}i_pq_{p-1}}' \mu_{r_{p}i_pr_{p-1}}'\mu_{q_{p}j_p'q_{p-1}}\mu_{r_{p}i_p'r_{p-1}}k(\xb_{i_p}^p,\xb_{i_p'}^p)k(\xb_{i_p}^p,\xb_{j_p'}^p)\\&+\sum_{\substack{r_1,...,r_P\\i_1',...,i_P'\\j_1',...,j_P'}} \prod_{p=1}^P\sigma_{r_{p}i_pr_{p-1}}'^2\sigma_{i_p'}\sigma_{j_p'}k(\xb_{i_p}^p,\xb_{i_p'}^p)k(\xb_{i_p}^p,\xb_{j_p'}^p)\\&+\sum_{\substack{r_1,...,r_P\\i_1',...,i_P'\\j_1',...,j_P'}}\prod_{p=1}^P \sigma_{r_{p}i_pr_{p-1}}'^2\mu_{r_{p}i_p'r_{p-1}}\mu_{r_{p}j_p'r_{p-1}}k(\xb_{i_p}^p,\xb_{i_p'}^p)k(\xb_{i_p}^p,\xb_{j_p'}^p)\\&+\sum_{\substack{r_1,...,r_P\\i_1',...,i_P'\\j_1',...,j_P'}}\prod_{p=1}^P \sigma_{i_p'}\sigma_{j_p'}\mu_{r_{p}i_pr_{p-1}}'^2k(\xb_{i_p}^p,\xb_{i_p'}^p)k(\xb_{i_p}^p,\xb_{j_p'}^p)
\end{split}
\end{equation}
\begin{equation}
    \begin{split}
        &\Longleftrightarrow \Bigg(\prod_p^P \times_{-1}   \Mb_p' \circ (\Mb_p\times_{2} \Kb^p)\Bigg)^2 + \Bigg(\prod_{p}^P\times_{-1}\Sigma_p'\circ ( \mathbf{1}\times_2  \Big(\Ib\cdot(\Kb^p \cdot \Bb_p)^2\cdot \bar{1})\Big) \Bigg)\\&+\Bigg(\prod_p^P \times_{-1}(\Sigma_p' \circ(\Mb_p\times_2 \Kb^p)^2) \Bigg)+ \Bigg(\prod_{p}^P\times_{-1} \Mb_p'^2\circ ( \mathbf{1}\times_2 \Big(\Ib\cdot(\Kb^p \cdot \Bb_p)^2\cdot \bar{1})\Big) \Bigg)
    \end{split}
\end{equation}
where $\Sigma_p =  \Bb_p  \Bb_p^T$, $\mathbf{1}$ denotes a constant one tensor with the same dimensions as $\Sigma_p'$, $\hat{1}\in \RR^{1\times R}$ where $R$ is the column dimension of $\Bb_p$, $\Ib$ the identity matrix and $\Sigma_p'$ is the same as in the univariate case.  For RFF's we have that
\begin{equation}
    (\Kb^p \cdot \Bb_p)^2\cdot \bar{1}\approx ((\Phi_p\cdot\Phi_p^{\top}) \cdot \Bb_p)^2 \cdot \bar{1} + \text{vec}(D_p^2)= (\Phi_p\cdot(\Phi_p^{\top} \cdot \Bb_p))^2\cdot \bar{1}+ \text{vec}(D_p^2).
\end{equation}
As the middle term remains the same as in the univariate case our full reconstruction term is
 
 \begin{equation}
    \begin{split}
        &\mathbb{E}_{q}[\log{p(y_{i_1,...,i_P}|\Vb_1,...,\Vb_p,\Vb_1',...,\Vb_p')}] \propto \\& \frac{1}{\sigma_y^2} \Bigg[\Yb^2-2\Yb \circ \Bigg( \prod_{p=1}^P \times_{-1} ( \Mb_p' \circ (\Mb_p\times_{2} \Kb^p))\Bigg) + \Bigg(\prod_p^P \times_{-1}   \Mb_p' \circ (\Mb_p\times_{2} \Kb^p)\Bigg)^2 \\&+ \Bigg(\prod_{p}^P\times_{-1}\Sigma_p'\circ ( \mathbf{1}\times_2  \Big(\Ib\cdot(\Kb^p \cdot \Bb_p)^2\cdot \bar{1})\Big) \Bigg)\\&+\Bigg(\prod_p^P \times_{-1}(\Sigma_p' \circ(\Mb_p\times_2 \Kb^p)^2) \Bigg)+ \Bigg(\prod_{p}^P\times_{-1} \Mb_p'^2\circ ( \mathbf{1}\times_2 \Big(\Ib\cdot(\Kb^p \cdot \Bb_p)^2\cdot \bar{1})\Big) \Bigg]
    \end{split}
\end{equation}
 
\subsubsection{Latent Scaling}
Assuming that $\sigma_y$ is a constant hyperparameter. We first have that 
\begin{equation}
\begin{split}
    &\mathbb{E}_{q}[\log{p(y_{i_1,...,i_P}|\Vb_1,...,\Vb_p,\Vb_1',...,\Vb_p')}] \\ &\propto \frac{1}{\sigma_y^2} \mathbb{E}_{q}[(y_{i_1,...,i_P}- \big(\sum_{s_1,...,s_P}\prod_{p=1}^P v_{s_pi_ps_{p-1}}^s\sum_{\substack{r_1,...,r_P\\i_1',...,i_P'}} \prod_{p=1}^P   v_{r_{p}i_p'r_{p-1}}k(\xb_{i_p}^p,\xb_{i_p'}^p)+\sum_{b_1,...,b_P}\prod_{p=1}^P v_{b_pi_pb_{p-1}}^b\big) )^2 ] \\ &= \frac{1}{\sigma_y^2} \Bigg(y_{i_1,...,i_P}^2 - 2y_{i_1,...,i_P}\cdot \big(\underbrace{\sum_{s_1,...,s_P}\prod_{p=1}^P \mathbb{E}_{q}[v_{s_pi_ps_{p-1}}^s]}_{f_{v_{s_pi_ps_{p-1}}^s}}   \underbrace{\sum_{\substack{r_1,...,r_P\\i_1',...,i_P'}} \prod_{p=1}^P   \mathbb{E}_{q}[v_{r_{p}i_p'r_{p-1}}]k(\xb_{i_p}^p,\xb_{i_p'}^p)}_{f_{v_{r_{p}i_p'r_{p-1}}}}\\&+\underbrace{\sum_{b_1,...,b_P}\prod_{p=1}^P \mathbb{E}_{q}[v_{b_pi_pb_{p-1}}^b]}_{f_{v_{b_pi_pb_{p-1}}^b}}\big) +\sum_{\substack{s_1,...,s_P\\s_1',...,s_P'}}\prod_{p=1}^P \mathbb{E}_{q}[v_{s_pi_ps_{p-1}}^sv_{s_p'i_ps_{p-1}'}^s]\\&\sum_{\substack{r_1,...,r_P\\q_1,...,q_P\\i_1',...,i_P'\\j_1',...,j_P'}} \prod_{p=1}^P   \mathbb{E}_{q}[v_{r_{p}i_p'r_{p-1}}v_{q_{p}j_p'q_{p-1}}]k(\xb_{i_p}^p,\xb_{i_p'}^p)k(\xb_{i_p}^p,\xb_{j_p'}^p) \\&+2f_{v_{s_pi_ps_{p-1}}^s}f_{v_{r_{p}i_p'r_{p-1}}}f_{v_{b_pi_pb_{p-1}}^b}+\sum_{\substack{b_1,...,b_P\\b_1',...,b_P'}}\prod_{p=1}^P \mathbb{E}_{q}[v_{b_pi_pb_{p-1}}^b v_{b_p'i_pb_{p-1}'}^b]  \Bigg).
\end{split}
\end{equation} 

\paragraph{Univariate case}
For the univariate case, the squared component of the reconstruction term in the ELBO becomes
\begin{equation}
\begin{split}
    &\Bigg((\prod_{p=1}^P \times_{-1} \Mb_p^s)^2+(\prod_{p=1}^P \times_{-1} \Sigma_p^s) \Bigg)\circ\Bigg((\prod_p^P \times_{-1} \Mb_p\times_{2} \Kb^p)^2+ \prod_{p=1}^P \times_{-1} \Sigma_p\times_{2} (\Kb^p)^2\Bigg)\\&+2(\prod_{p=1}^P \times_{-1} \Mb_p^s)\circ(\prod_{p=1}^P \times_{-1} \Mb_p^b)\circ(\prod_{p=1}^P \times_{-1} \Mb_p \times_2 \Kb^p) +\Bigg((\prod_{p=1}^P \times_{-1} \Mb_p^b)^2+(\prod_{p=1}^P \times_{-1} \Sigma_p^b) \Bigg).
\end{split}
\end{equation} 
Too see this, we notice that all of the sums from the weighted latent regression case reappears here as well. The entire expression is given by

\begin{equation}
\begin{split}
     &\mathbb{E}_{q}[\log{p(y_{i_1,...,i_P}|\Vb_1,...,\Vb_p,\Vb_1',...,\Vb_p')}] \propto \\&\frac{1}{\sigma_y^2} \Bigg[\Yb^2 -2\Yb\circ\Bigg(\prod_{p=1}^P (\Mb_p^s \circ (\Mb_p \times_2 \Kb^p) + \Mb_p^b) \Bigg)
    \\&+\Bigg((\prod_{p=1}^P \times_{-1} \Mb_p^s)^2+(\prod_{p=1}^P \times_{-1} \Sigma_p^s) \Bigg)\circ\Bigg((\prod_p^P \times_{-1} \Mb_p\times_{2} \Kb^p)^2+ \prod_{p=1}^P \times_{-1} \Sigma_p\times_{2} (\Kb^p)^2\Bigg)\\&+2(\prod_{p=1}^P \times_{-1} \Mb_p^s)\circ(\prod_{p=1}^P \times_{-1} \Mb_p^b)\circ(\prod_{p=1}^P \times_{-1} \Mb_p \times_2 \Kb^p) +\Bigg((\prod_{p=1}^P \times_{-1} \Mb_p^b)^2+(\prod_{p=1}^P \times_{-1} \Sigma_p^b) \Bigg)\Bigg].
\end{split}
\end{equation}

\paragraph{Multivariate case}
For multivariate case
\begin{equation}
\begin{split}
    &\Bigg((\prod_{p=1}^P \times_{-1} \Mb_p^s)^2+(\prod_{p=1}^P \times_{-1} \Sigma_p^s) \Bigg)\circ\Bigg((\prod_p^P \times_{-1} \Mb_p\times_{2} \Kb^p)^2+ \prod_{p=1}^P \times_{-1} (\mathbf{1}\times_2\Big(\Ib\cdot(\Kb^p \cdot \Bb_p)^2\cdot \bar{1})\Big)\Bigg)\\&+2(\prod_{p=1}^P \times_{-1} \Mb_p^s)\circ(\prod_{p=1}^P \times_{-1} \Mb_p^b)\circ(\prod_{p=1}^P \times_{-1} \Mb_p \times_2 \Kb^p)+\Bigg((\prod_{p=1}^P \times_{-1} \Mb_p^b)^2+(\prod_{p=1}^P \times_{-1} \Sigma_p^b) \Bigg)
\end{split}
\end{equation}
Again, we notice that all of the sums from the weighted latent regression case reappears here as well. The full expression is given by
\begin{equation}
\begin{split}
     &\mathbb{E}_{q}[\log{p(y_{i_1,...,i_P}|\Vb_1,...,\Vb_p,\Vb_1',...,\Vb_p')}] \propto \\&\frac{1}{\sigma_y^2} \Bigg[\Yb^2 -2\Yb\circ\Bigg(\prod_{p=1}^P (\Mb_p^s \circ (\Mb_p \times_2 \Kb^p) + \Mb_p^b) \Bigg)
    \\+&\Bigg((\prod_{p=1}^P \times_{-1} \Mb_p^s)^2+(\prod_{p=1}^P \times_{-1} \Sigma_p^s) \Bigg)\circ\Bigg((\prod_p^P \times_{-1} \Mb_p\times_{2} \Kb^p)^2+ \prod_{p=1}^P \times_{-1} (\mathbf{1}\times_2\Big(\Ib\cdot(\Kb^p \cdot \Bb_p)^2\cdot \bar{1})\Big)\Bigg)\\&+2(\prod_{p=1}^P \times_{-1} \Mb_p^s)\circ(\prod_{p=1}^P \times_{-1} \Mb_p^b)\circ(\prod_{p=1}^P \times_{-1} \Mb_p \times_2 \Kb^p)+\Bigg((\prod_{p=1}^P \times_{-1} \Mb_p^b)^2+(\prod_{p=1}^P \times_{-1} \Sigma_p^b) \Bigg) \Bigg]
\end{split}
\end{equation}


%
%


\end{document}